\documentclass[runningheads]{llncs}
\usepackage{graphicx}
\usepackage{amsmath,amssymb} 
\usepackage{epsfig}
\usepackage{epstopdf}
\usepackage{color}
\usepackage[colorlinks,linkcolor=red,anchorcolor=blue,citecolor=green]{hyperref}
\usepackage[width=122mm,left=12mm,paperwidth=146mm,height=193mm,top=12mm,paperheight=217mm]{geometry}
\usepackage{graphicx}
\usepackage{subfigure}
\usepackage{ulem}
\usepackage{multirow}
\usepackage{makecell}
\usepackage{cite}
\usepackage{bbding}

\begin{document}
\pagestyle{headings}
\mainmatter
\def\ECCV18SubNumber{399}  

\title{Learning Warped Guidance for Blind Face Restoration} 

\titlerunning{}
\authorrunning{X. Li et al.}

\author{Xiaoming Li$^1$, Ming Liu$^1$, Yuting Ye$^1$, Wangmeng Zuo$^{1(}$\Envelope$^)$, Liang Lin$^2$, \newline and Ruigang Yang$^3$}
\institute{$^1$School of Computer Science and Technology, Harbin Institute of Technology, China \newline \email{csxmli@hit.edu.cn}, \email{csmliu@outlook.com}, \email{yeyuting.jlu@gmail.com}, \email{wmzuo@hit.edu.cn} \newline $^2$School of Data and Computer Science, Sun Yat-sen University, China\newline \email{linliang@ieee.org} \newline $^3$Department of Computer Science, University of Kentucky, USA \newline \email{ryang@cs.uky.edu}}
\maketitle
\begin{abstract}
This paper studies the problem of blind face restoration from an unconstrained blurry, noisy, low-resolution, or compressed image (i.e., degraded observation).
For better recovery of fine facial details, we modify the problem setting by taking both the degraded observation and a high-quality guided image of the same identity as input to our guided face restoration network (GFRNet).
However, the degraded observation and guided image generally are different in pose, illumination and expression, thereby making plain CNNs (e.g., U-Net~\cite{ronneberger2015u}) fail to recover fine and identity-aware facial details.
To tackle this issue, our GFRNet model includes both a warping subnetwork (WarpNet) and a reconstruction subnetwork (RecNet).
The WarpNet is introduced to predict flow field for warping the guided image to correct pose and expression (i.e., warped guidance),
while the RecNet takes the degraded observation and warped guidance as input to produce the restoration result.
Due to that the ground-truth flow field is unavailable, landmark loss together with total variation regularization are incorporated to guide the learning of WarpNet.
Furthermore, to make the model applicable to blind restoration, our GFRNet is trained on the synthetic data with versatile settings on blur kernel, noise level, downsampling scale factor, and JPEG quality factor.
Experiments show that our GFRNet not only performs favorably against the state-of-the-art image and face restoration methods, but also generates visually photo-realistic results on real degraded facial images.

\end{abstract}
\section{Introduction}\label{section1}
Face restoration aims to reconstruct high quality face image from degraded observation for better display and further analysis~\cite{baker2000hallucinating,liu2007face,zhu2016deep,yu2016ultra,cao2017attention,chen2017fsrnet,yu2017face,huang2017wavelet,xu2017learning,yu2017hallucinating,chrysos2017deep}.
In the ubiquitous imaging era, imaging sensors are embedded into many consumer products and surveillance devices, and more and more images are acquired under unconstrained scenarios.
Consequently, low quality face images cannot be completely avoided during acquisition and communication due to the introduction of low-resolution, defocus, noise and compression.
On the other hand, high quality face images are sorely needed for human perception, face recognition~\cite{phillips2005overview} and other face analysis~\cite{andreu2014analysis} tasks.
All these make face restoration a very challenging yet active research topic in computer vision.

\begin{figure}[t]
\centering
\subfigure[]{
  \begin{minipage}[b]{.16\columnwidth}
    \includegraphics[width=0.99\textwidth]{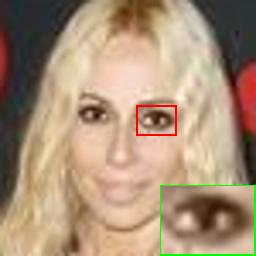}\\
    \includegraphics[width=0.99\textwidth]{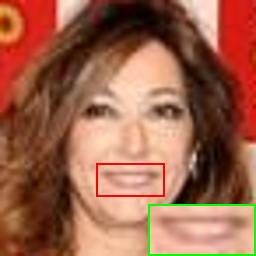}
  \end{minipage}
}
\hspace{-3ex}
\subfigure[]{
  \begin{minipage}[b]{.16\columnwidth}
    \includegraphics[width=0.99\textwidth]{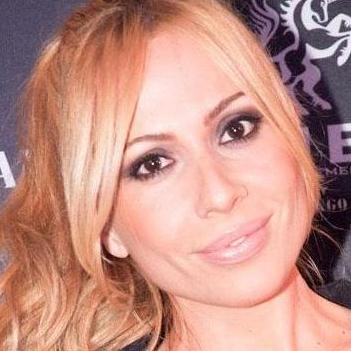}\\
    \includegraphics[width=0.99\textwidth]{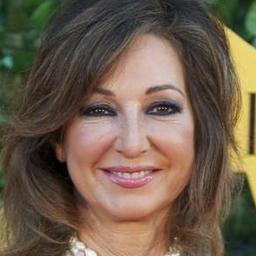}
  \end{minipage}
}
\hspace{-3ex}
\subfigure[]{
  \begin{minipage}[b]{.16\columnwidth}
    \includegraphics[width=0.99\textwidth]{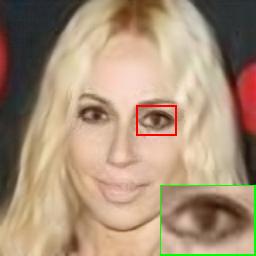}\\
    \includegraphics[width=0.99\textwidth]{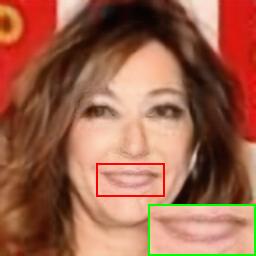}
  \end{minipage}
}
\hspace{-3ex}
\subfigure[]{
  \begin{minipage}[b]{.16\columnwidth}
    \includegraphics[width=0.99\textwidth]{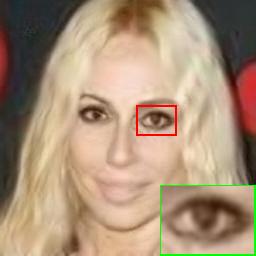}\\
    \includegraphics[width=0.99\textwidth]{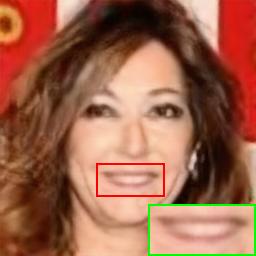}
  \end{minipage}
}
\hspace{-3ex}
\subfigure[]{
  \begin{minipage}[b]{.16\columnwidth}
    \includegraphics[width=0.99\textwidth]{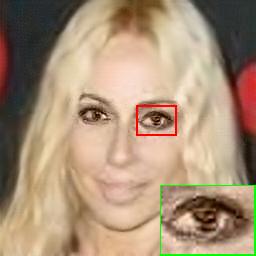}\\
    \includegraphics[width=0.99\textwidth]{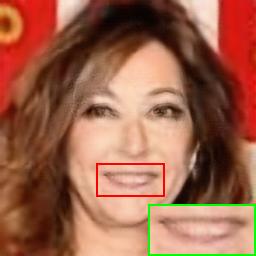}
  \end{minipage}
}
\hspace{-3ex}
\subfigure[]{
  \begin{minipage}[b]{.16\columnwidth}
    \includegraphics[width=0.99\textwidth]{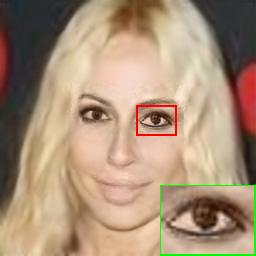}\\
    \includegraphics[width=0.99\textwidth]{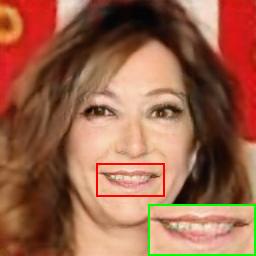}
  \end{minipage}
}
  \caption{Restoration results on real low quality images: (a)~real low quality image, (b)~guided image, and the results by (c)~U-Net~\cite{ronneberger2015u} by taking low quality image as input, (d)~U-Net~\cite{ronneberger2015u} by taking both guided image and low quality image as input, (e)~our GFRNet without landmark loss, and (f)~our full GFRNet model. Best viewed by zooming in the screen.}
  \label{fig:realfig1}
\end{figure}

Many studies have been carried out to handle specific face restoration tasks, such as denoising~\cite{anwar2017category,Anwar2017BMVC}, hallucination~\cite{baker2000hallucinating,liu2007face,zhu2016deep,yu2016ultra,cao2017attention,chen2017fsrnet,yu2017face,huang2017wavelet,xu2017learning,yu2017hallucinating} and deblurring \cite{chrysos2017deep}.
Most existing methods, however, are proposed for handling a single specific face restoration task in a non-blind manner.
In practical scenario, it is more general that both the degradation types and degradation parameters are unknown in advance.
Therefore, more attentions should be given to blind face restoration.
Moreover, most previous works produce the restoration results purely relying on a single degraded observation.
It is worth noting that the degradation process generally is highly ill-posed.
By learning a direct mapping from degraded observation, the restoration result inclines to be over-smoothing and cannot faithfully retain fine and identity-aware facial details.

In this paper, we study the problem of guided blind face restoration by incorporating the degraded observation and a high-quality guided face image.
Without loss of generality, the guided image is assumed to have the same identity with the degraded observation, and is frontal with eyes open. %
We note that such guided restoration setting is practically feasible in many real world applications.
For example, most smartphones support to recognize and group the face images according to their identities\footnote{https://support.apple.com/HT207103}.
In each group, the high quality face image can thus be exploited to guide the restoration of low quality images.
In film restoration, it is also encouraging to use the high quality portrait of an actor to guide the restoration of low-resolution and corrupted face images of the same actor from an old film.
For these tasks,
further incorporation of guided image not only can ease the difficulty of blind restoration, but also is helpful in faithfully recovering fine and identity-aware facial details.

Guided blind face restoration, however, cannot be addressed well by simply taking the degraded observation and guided image as input to plain convolutional networks (CNNs), due to that the two images generally are of different poses, expressions and lighting conditions.
Fig.~\ref{fig:realfig1}(c) shows the results obtained using the U-Net~\cite{ronneberger2015u} by only taking degraded observation as input, while Fig.~\ref{fig:realfig1}(d) shows the results by taking both two images as input.
It can be seen that direct incorporation of guided image brings very limited improvement on the restoration result.
To tackle this issue, we develop a guided {face} restoration network (GFRNet) consisting of a warping subnetwork (WarpNet) and a reconstruction subnetwork (RecNet).
Here, the WarpNet is firstly deployed to predict a flow field for warping the guided image to obtain the warped guidance, which is required to have the same pose and expression with degraded observation.
Then, the RecNet takes both degraded observation and warped guidance as input to produce the final restoration result.
To train GFRNet, we adopt the reconstruction learning to constrain the restoration result to be close to the target image (i.e., ground-truth), and further employ the adversarial learning for visually realistic restoration.

Nonetheless, even though the WarpNet can be end-to-end trained with reconstruction and adversarial learning, we empirically find that it cannot converge to the desired solution and fails to align the guided image to the correct pose and expression.
Fig.~\ref{fig:realfig1}(e) gives the results of our GFRNet trained by reconstruction and adversarial learning.
One can see that its improvement over U-Net is still limited, especially when the degraded observation and guided images are distinctly different in pose.
Moreover, the ground-truth flow field is unavailable, and the target and guided images may be of different lighting conditions, making it infeasible to directly use the target image to guide the WarpNet learning.
Instead, we adopt the face alignment method~\cite{TCDCN} to detect the face landmarks of the target and guided images, and then introduce the landmark loss as well as the total variation (TV) regularizer to train the WarpNet.
{As in} Fig.~\ref{fig:realfig1}(f), our full GFRNet achieves the favorable visual quality, and is effective in recovering fine facial details.
Furthermore, to make the learned GFRNet applicable to blind face restoration, our model is trained on the synthetic data generated by a general degradation model with versatile settings on blur kernel, noise level, downsampling scale factor, and JPEG quality factor.

Extensive experiments are conducted to evaluate the proposed GFRNet for guided blind face restoration.
The results show that the WarpNet is effective in aligning the guided image to the desired pose and expression.
The proposed GFRNet achieves significant performance gains over the state-of-the-art restoration methods, e.g., SRCNN~\cite{dong2014learning}, VDSR~\cite{kim2016accurate}, SRGAN~\cite{Ledig2017CVPR}, DCP~\cite{pan2016blind}, {DeepDeblur~\cite{Nah2017CVPR}, DeblurGAN~\cite{DeblurGAN}, DnCNN~\cite{zhang2017beyond}, MemNet~\cite{MemNet}, ARCNN~\cite{Dong2015ICCV}, CBN~\cite{zhu2016deep}, WaveletSRNet~\cite{huang2017wavelet}, TDAE~\cite{yu2017hallucinating},} SCGAN~\cite{xu2017learning} and MCGAN~\cite{xu2017learning} in terms of both quantitative metrics (i.e., PSNR and SSIM) and visually perceptual quality.
Moreover, our GFRNet also performs favorably on real degraded images as shown in Fig.~\ref{fig:realfig1}(f).
To sum up, the main contribution of this work includes:
\begin{itemize}
\item The GFRNet architecture for guided blind face restoration, which includes a warping subnetwork (WarpNet) and a reconstruction subnetwork (RecNet).
\item The incorporation of landmark loss and TV regularization for training the WarpNet.
\item The promising results of GFRNet on both synthetic and real face images.
\end{itemize}

\section{Related Work}\label{section2}
Recent years have witnessed the unprecedented success of deep learning in many image restoration tasks such as super-resolution~\cite{dong2014learning,kim2016accurate,Ledig2017CVPR}, {denoising~\cite{zhang2017beyond,MemNet}}, {compression artifact removal~\cite{Dong2015ICCV,Galteri2017ICCV}}, compressed sensing~\cite{kulkarni2016reconnet,jin2017deep,lucas2018using}, and {deblurring \cite{Nah2017CVPR,DeblurGAN,chakrabarti2016neural,nimisha2017blur}}.
As to face images, several CNN architectures have been developed for face hallucination~\cite{zhu2016deep,cao2017attention,chen2017fsrnet,huang2017wavelet}, and the adversarial learning is also introduced to enhance the visual quality~\cite{yu2016ultra,yu2017face}.
{Most of these methods}, however, are suggested for non-blind restoration and are restricted by the specialized tasks.
Benefitted from the powerful modeling capability of deep CNNs, recent studies have shown that it is feasible to train a single model for handling multiple instantiations of degradation (e.g., different noise levels)~\cite{zhang2017beyond,mao2016image}.
As for face hallucination, Yu et al.~\cite{yu2017face,yu2017hallucinating} suggest one kind of transformative discriminative networks to super-resolve different unaligned tiny face images.
Nevertheless, blind restoration is a more challenging problem and requires to learn a single model for handling all instantiations of one or more degradation types.

Most studies on deep blind restoration are given to blind deblurring, which aims to recover the latent clean image from noisy and blurry observation with unknown degradation parameters.
Early learning-based or CNN-based blind deblurring methods~\cite{chakrabarti2016neural,schuler2016learning,xiao2016learning} usually follow traditional framework which includes a blur kernel estimation stage and a non-blind deblurring stage.
With the rapid progress and powerful modeling capability of CNNs, recent studies incline to bypass blur kernel estimation by directly training a deep model to restore clean image from degraded observation~\cite{Nah2017CVPR,DeblurGAN,nimisha2017blur,noroozi2017motion,hradivs2015convolutional}.
As to blind face restoration, Chrysos and Zafeiriou~\cite{chrysos2017deep} utilize a modified ResNet architecture to perform face deblurring, while Xu et al.~\cite{xu2017learning} adopt the generative adversarial network (GAN) framework to super-resolve blurry face image.
It is worth noting that the success of such kernel-free end-to-end approaches depends on both the modeling capability of CNN and the sufficient sampling on clean images and degradation parameters, making it difficult to design and train.
Moreover, the highly ill-posed degradation further increases the difficulty of recovering the correct fine details only from degraded observation~\cite{lin2008limits}.
In this work, we elaborately tackle this issue by incorporating a high quality guided image and designing appropriate network architecture and learning objective.

Several learning-based and CNN-based approaches are also developed for color-guided depth image enhancement~\cite{li2016deep,hui2016depth,gu2017learning}, where the structural interdependency between intensity and depth image is modeled and exploited to reconstruct high quality depth image.
For guided depth image enhancement, Hui et al.~\cite{hui2016depth} present a CNN model to learn multi-scale guidance, while Gu et al.~\cite{gu2017learning} incorporate weighted analysis representation and truncated inference for dynamic guidance learning.
For general guided filtering, Li et al.~\cite{li2016deep} construct CNN-based joint filters to transfer structural details from guided image to reconstructed image.
However, these approaches assume that the guided image is spatially well aligned with the degraded observation.
Due to that the guided image and degraded observation usually are different in pose and expression, such assumption generally does not hold true for guided face restoration.
To address this issue, a WarpNet is introduced in our GFRNet to learn a flow field for warping the guided image to the desired pose and expression.

Recently, spatial transformer networks (STNs) are suggested to learn a spatial mapping for warping an image~\cite{jaderberg2015spatial}, and appearance flow networks (AFNs) are presented to predict a dense flow field to move pixels \cite{zhou2016view,ganin2016deepwarp}.
Deep dense flow networks have been applied to view synthesis~\cite{zhou2016view,park2017cvpr}, gaze manipulation~\cite{ganin2016deepwarp}, expression editing~\cite{yeh2016semantic}, and video frame synthesis~\cite{liu2017iccv}.
In these approaches, the target image is required to have the similar lighting condition with the input image to be warped, and the dense flow networks can thus be trained via reconstruction learning.
However, in our guided face restoration task, the guided image and the target image usually are of different lighting conditions, making it less effective to train the flow network via reconstruction learning.
Moreover, the ground-truth dense flow field is not available, further increasing the difficulty to train WarpNet.
To tackle this issue, we use the face alignment method~\cite{TCDCN} to extract the face landmarks of guided and target images.
Then, the landmark loss and TV regularization are incorporated to facilitate the WarpNet training.
\section{Proposed Method}\label{section3}
This section presents our GFRNet to recover high quality face image from degraded observation with unknown degradation. 
Given a degraded observation $I^d$ and a guided image $I^g$, our GFRNet model produces the restoration result $\hat{I} = \mathcal{F}(I^d, I^g)$ to approximate the ground-truth target image $I$.
Without loss of generality, $I^g$ and $I$ are of the same identity and image size $256 \times 256$.
Moreover, to provide richer guidance information, $I^g$ is assumed to be of high quality, frontal, non-occluded with eyes open.
Nonetheless, we empirically find that our GFRNet is robust when the assumption is violated.
For simplicity, we also assume $I^d$ also has the same size with $I^g$.
When such assumption does not hold, e.g., in face hallucination, we simply apply the bicubic scheme to upsample $I^d$ to the size of $I^g$ before inputting it to the GFRNet.

In the following, we first describe the GFRNet model as well as the network architecture.
Then, a general degradation model is introduced to generate synthetic training data. 
Finally, we present the model objective of our GFRNet.
\subsection{Guided Face Restoration Network}
\begin{figure*}[t]
\setlength{\abovecaptionskip}{-1ex}
\setlength{\belowcaptionskip}{-2ex}
\begin{center}
\includegraphics[width=1\linewidth]{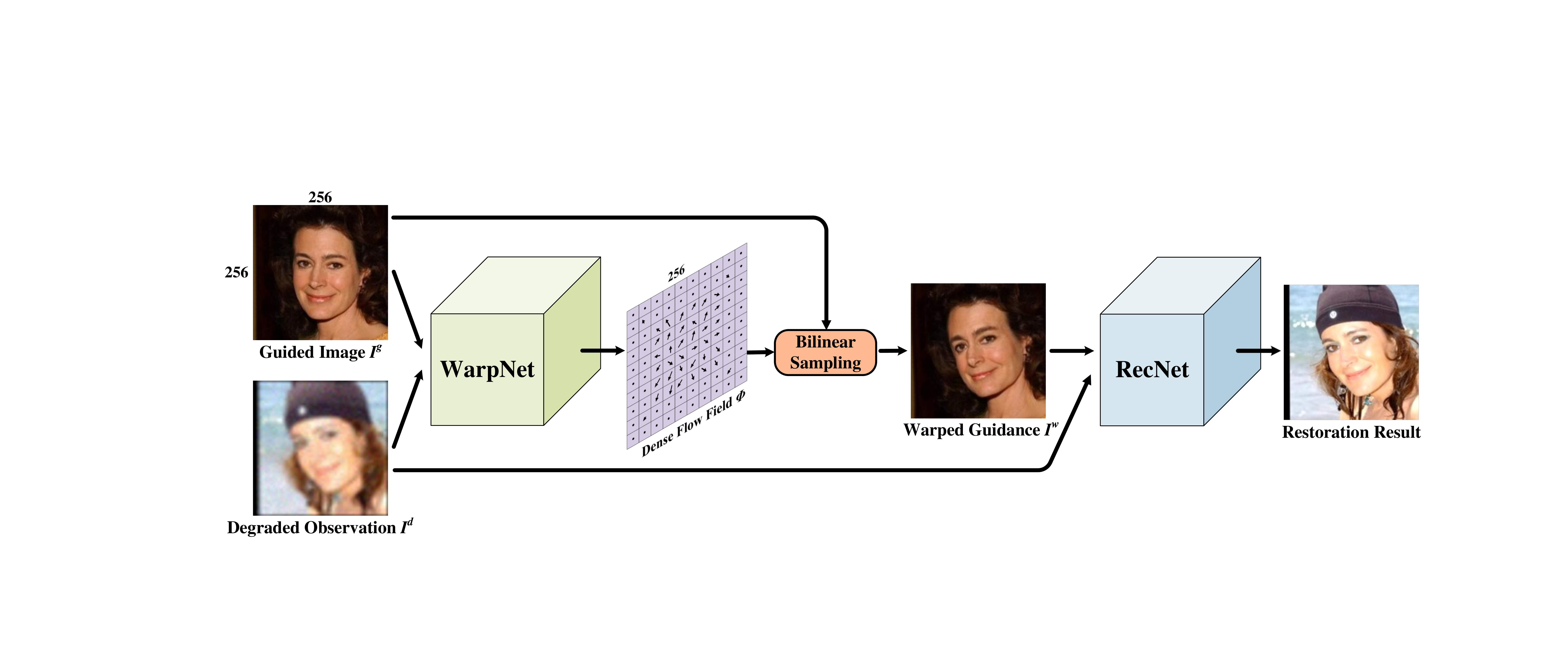}
\end{center}
   \caption{Overview of our GFRNet. The WarpNet takes the degraded observation $I^d$ and guided image $I^g$ as input to predict the dense flow field $\Phi$, which is adopted to deform $I^g$ to the warped guidance $I^w$. $I^w$ is expected to be spatially well aligned with ground-truth $I$. Thus the RecNet takes $I^w$ and $I^d$ as input to produce the restoration result $\hat{I}$. }
   \label{fig:pipeline}
\end{figure*}
The degraded observation $I^d$ and guided image $I^g$ usually vary in pose and expression.
{Directly} taking $I^d$ and $I^g$ as input to plain CNNs generally cannot achieve much performance gains over taking only $I^d$ as input (See Fig.~\ref{fig:realfig1}(c)(d)).
To address this issue, the proposed GFRNet consists of two subnetworks: (i) the warping subnetwork (WarpNet) and (ii) reconstruction subnetwork (RecNet).

Fig.~\ref{fig:pipeline} illustrates the overall architecture of our GFRNet.
The WarpNet takes $I^d$ and $I^g$ as input to predict the flow field for warping guided image,
\begin{equation}
\label{flow_field}
\Phi = \mathcal{F}_w(I^d, I^g; \Theta_w),
\end{equation}
where $\Theta_w$ denotes the WarpNet model parameters.
With $\Phi$, the output pixel value of the warped guidance $I^{w}$ at location $(i,j)$ is given by
\begin{equation}
\label{warped_guidance}
I^{w}_{i,j} = \sum_{(h,w) \in \mathcal{N}} I^g_{h,w} \max(0, 1 - |\Phi^{y}_{i,j} - h|) \max(0, 1 - |\Phi^{x}_{i,j} - w|),
\end{equation}
where $\Phi^{x}_{i,j}$ and $\Phi^{y}_{i,j}$ denote the predicted $x$ and $y$ coordinates for the pixel $I^{w}_{i,j}$, respectively.
$\mathcal{N}$ stands for the 4-pixel neighbors of $(\Phi^{x}_{i,j}, \Phi^{y}_{i,j})$.
From Eqn.~(\ref{warped_guidance}), we note that $I^{w}$ is subdifferentiable to $\Phi$~\cite{jaderberg2015spatial}.
Thus, the WarpNet can be end-to-end trained by minimizing the losses defined either on $I^{w}$ or on $\Phi$.

The predicted warping guidance $I^{w}$ is expected to have the same pose and expression with the ground-truth $I$.
Thus, the RecNet takes $I^d$ and $I^w$ as input to produce the final restoration result,
\begin{equation}
\label{reconstruction}
\hat{I} = \mathcal{F}_r(I^d, I^w; \Theta_r),
\end{equation}
where $\Theta_r$ denotes the RecNet model parameters.

\subsubsection{Warping Subnetwork (WarpNet).}
%
The WarpNet adopts the encoder-decoder structure and is comprised of two major components:
\begin{itemize}
\item The \emph{input encoder} extracts feature representation from $I^{d}$ and $I^g$, consisting of eight convolution layers and each one with size $4 \times 4$ and stride $2$.
\item The \emph{flow decoder} predicts the dense flow field for warping $I^g$ to the desired pose and expression, consisting of eight deconvolution layers.
\end{itemize}
\begin{figure*}[t]
\setlength{\abovecaptionskip}{-1ex}
\setlength{\belowcaptionskip}{-2ex}
\begin{center}
\includegraphics[width=1\linewidth]{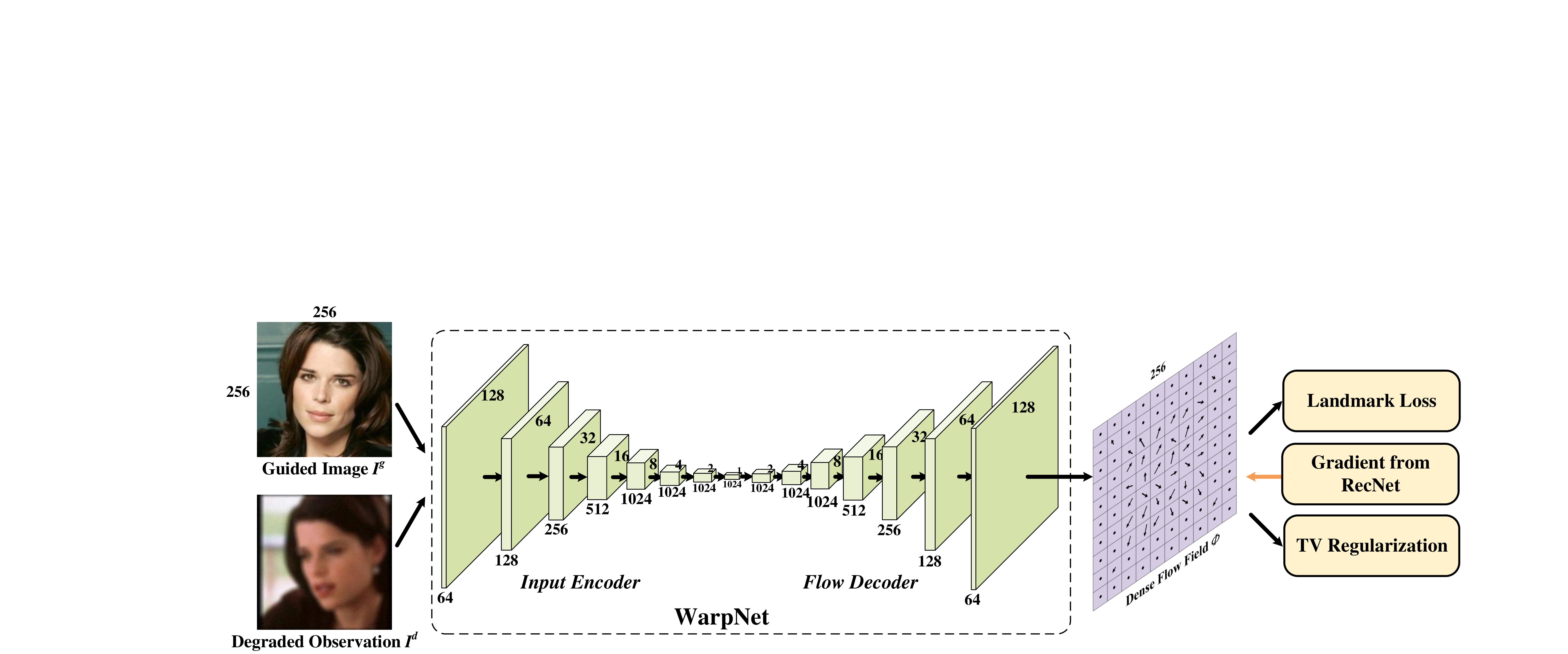}
\end{center}
   \caption{Architecture of our WarpNet. It takes the degraded observation $I^d$ and guided image $I^g$ as input to predict the dense flow field $\Phi$, which is adopted to deform $I^g$ to the warped guidance $I^w$. $I^w$ is expected to be spatially well aligned with ground-truth $I$. Landmark loss, TV regularization as well as gradient from RecNet are deployed to facilitate the learning of WarpNet.
    }
   \label{fig:flownet}
\end{figure*}
Except the first layer in encoder and the last layer in decoder, all the other layers adopt the convolution-BatchNorm-ReLU form. The detailed structure of WarpNet is shown in Fig~\ref{fig:flownet}.

Finally, some explanations are given to the design of WarpNet.
(i) Instead of the U-Net architecture, we adopt the standard encoder-decoder structure by removing the skip connections.
It is worth noting that the input to encoder is two color images $I^d$ and $I^g$ while the output of decoder is a dense flow field $\Phi$.
Due to the heterogeneity of the input and output, it is unappropriate to concatenate the encoder features to the corresponding decoder features as in U-Net.
(ii) It is also improper to directly output the warped guidance instead of the flow field.
$I^w$ is of different pose and expression with $I^g$, making the U-Net architecture still suffer from the heterogeneity issue.
Due to the effect of the bottleneck (i.e., the fully connected layer), the encoder-decoder structure inclines to produce over-smoothing $I^w$.
Instead of directly predicting $I^w$, predicting the dense flow field $\Phi$ usually results in realistic facial image with fine details.

\subsubsection{Reconstruction Subnetwork (RecNet).}

\begin{figure*}[htb]
\setlength{\abovecaptionskip}{-1ex}
\setlength{\belowcaptionskip}{-2ex}
\begin{center}
\includegraphics[width=1\linewidth]{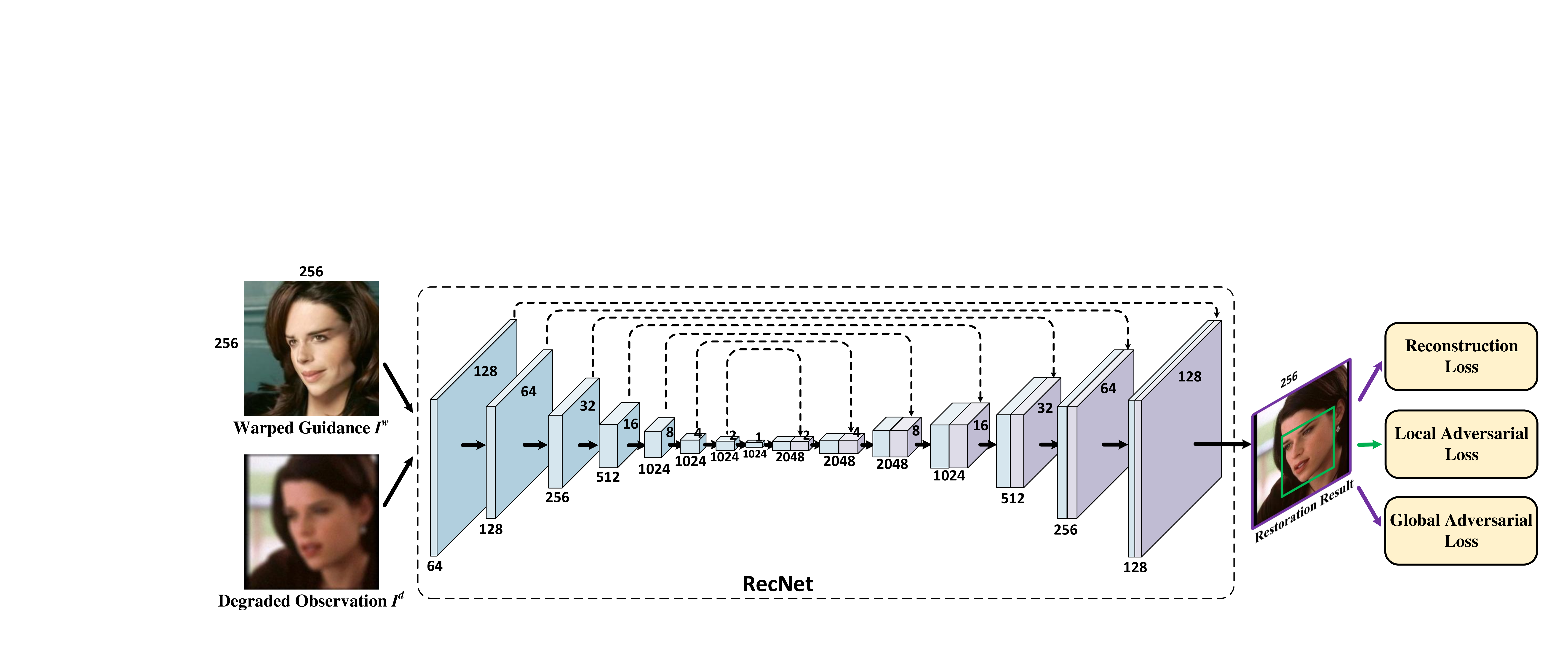}
\end{center}
   \caption{Architecture of our RecNet. It takes $I^w$ and $I^d$ as input to produce the restoration result $\hat{I}$. Reconstruction loss and global adversarial loss are adopted across entire image (labeled in \textcolor[RGB]{112,48,160}{\textbf{purple}}), while local adversarial loss is adopted across face region (labeled in \textcolor[RGB]{0,176,80}{\textbf{green}}).}
   \label{fig:recnet}
\end{figure*}

For the RecNet, the input ($I^d$ and $I^{w}$) are of the same pose and expression with the output ($\hat{I}$), and thus the U-Net can be adopted to produce the final restoration result $\hat{I}$.
The RecNet also includes two components, i.e., an encoder and a decoder.
The encoder and decoder of RecNet are of the same structure with those adopted in WarpNet.
To circumvent the information loss, the $i$-{th} layer is concatenated to the $(L-i)$-{th} layer via skip connections ($L$ is the depth of the U-Net), which has been demonstrated to benefit the rich and fine details of the generated image~\cite{pix2pix2016}. The detailed structure of RecNet is shown in Fig~\ref{fig:recnet}.
\subsection{Degradation Model and Synthetic Training Data}
To train our GFRNet, a degradation model is required to generate realistic degraded images.
We note that real low quality images can be the results of either defocus, long-distance sensing, noise, compression, or their combinations.
Thus, we adopt a general degradation model to generate degraded image $I^{d,s}$,
\begin{equation}
\label{eqn:degradation}
I^{d,s} = {\left( {\left( {I \otimes \mathbf{k}_{\varrho}} \right){ \downarrow _s} + \mathbf{n}_{\sigma}} \right)_{JPEG_q}},
\end{equation}
where $\otimes$ denotes the convolution operator.
$\mathbf{k}_{\varrho}$ stands for the Gaussian blur kernel with the standard deviation ${\varrho}$.
$\downarrow _s$ denotes the downsampling operator with scale factor $s$.
$\mathbf{n}_{\sigma}$ denotes the additive white Gaussian noise (AWGN) with the noise level $\sigma$.
$(\cdot)_{JPEG_q}$ denotes the JPEG compression operator with quality factor $q$.

In our general degradation model, ${\left( {I \otimes \mathbf{k}_{\varrho}} \right){ \downarrow _s} + \mathbf{n}_{\sigma}}$ characterizes the degradation caused by long-distance acquisition, while $(\cdot)_{JPEG_q}$ depicts the degradation caused by JPEG compression.
We also note that Xu et al.~\cite{xu2017learning} adopt the degradation model  ${\left( {I \otimes \mathbf{k}_{\varrho}}  + \mathbf{n}_{\sigma} \right){ \downarrow_s}}$.
However, to better simulate the long-distance image acquisition, it is more appropriate to add the AWGN on the downsampled image.
When $s \neq 1$, the degraded image $I^{d,s}$ is of different size with the ground-truth $I$.
So we use bicubic interpolation to upsample $I^{d,s}$ with scale factor $s$, and then take $I^d = (I^{d,s})\uparrow_s$ and $I^g$ as input to our GFRNet.

In the following, we explain the parameter settings for these operations:
\begin{itemize}
 \item \textbf{Blur kernel.} In this work, only the isotropic Gaussian blur kernel $\mathbf{k}_{\varrho}$ is considered to model the defocus effect.
     We sample the standard deviation of Gaussian blur kernel from the set $\varrho \in \{0, 1:0.1:3\}$.
 \item \textbf{Downsampler.} We adopt the bicubic downsampler as~\cite{zhu2016deep,cao2017attention,chen2017fsrnet,huang2017wavelet,xu2017learning}. The scale factor $s$ is sampled from the set $s \in \{1:0.1:8\}$.
 \item \textbf{Noise.} As for the noise level $\sigma$, we adopt the set $\sigma \in \{0:1:7\}$~\cite{xu2017learning}.
 \item \textbf{JPEG compression.} For economic storage and communication, JPEG compression with quality factor $q$ is further operated on the degraded image, and we sample $q$ from the set $q \in \{0, 10:1:40\}$. When $q = 0$, the image is only losslessly compressed.
\end{itemize}
By including $\varrho = 0$, $s=1$, $\sigma=0$ and $q=0$ in the set of degradation parameters, the general degradation model can simulate the effect of either the defocus, long-distance acquisition, noising, compression or their versatile combinations.

Given a ground-truth image $I_i$ together with the guided image $I^g_i$, we can first sample $\varrho_i$, $s_i$, $\sigma_i$ and $q_i$ from the parameter set, and then use the degradation model to generate a degraded observation $I^d_i$.
Furthermore, the face alignment method~\cite{TCDCN} is adopted to extract the landmarks $\{(x^{I_i}_{j}, y^{I_i}_{j})|_{j=1}^{68}\}$ for $I_i$ and $\{(x^{I^g_i}_{j}, y^{I^g_i}_{j})|_{j=1}^{68}\}$ for $I^g_i$.
Therefore, we define the synthetic training set as $\mathcal{X} = \{ (I_i, I^g_i, I^d_i, \{(x^{I_i}_{j}, y^{I_i}_{j})|_{j=1}^{68}\}, \{(x^{I^g_i}_{j}, y^{I^g_i}_{j})|_{j=1}^{68}\})|_{i=1}^N \}$, where $N$ denotes the number of samples.

\subsection{Model Objective}
\subsubsection{Losses on Restoration Result $\hat{I}$.}
To train our GFDNet, we define the reconstruction loss on the restoration result $\hat{I}$, and the adversarial loss is further incorporated on $\hat{I}$ to improve the visual perception quality.

\textbf{Reconstruction loss}.
The reconstruct loss is used to constrain the restoration result $\hat{I}$ to be close to the ground-truth $I$, which includes two terms.
The first term is the $\ell_2$ loss defined as the squared Euclidean distance between $\hat{I}$ and $I$, i.e., $ \ell_r^{0}(I, \hat{I}) = \| I - \hat{I} \|^2$.
Due to the inherent irreversibility of image restoration, only the $\ell_2$ loss inclines to cause over-smoothing result.
Following~\cite{johnson2016perceptual}, we define the second term as the perceptual loss on the pre-trained VGG-Face~\cite{parkhi2015deep}.
Denote by $\psi$ the VGG-Face network, $\psi_l(I)$ the feature map of the $l$-th convolution layer.
The perceptual loss on the $l$-th layer (i.e., Conv-4 in this work) is defined as
\begin{equation}\label{eqn:perceptual}
\ell_p^{\psi,l}(I, \hat{I}) = \frac{1}{C_l H_l W_l} \left\| {{\psi _l}(\hat{I}) - {\psi _l}(I)} \right\|_2^2
\end{equation}
where $C_l$, $H_l$ and $W_l$ denote the channel numbers, height and width of the feature map, respectively.
Finally, we define the reconstruction loss as
\begin{equation}\label{eqn:reconstruction}
\mathcal{L}_r(I, \hat{I}) = \lambda_{r,0} \ell_r^{0}(I, \hat{I}) + \lambda_{r,l} \ell_p^{\psi,l}(I, \hat{I}),
\end{equation}
where $\lambda_{r,0}$ and $\lambda_{r,l}$ are the tradeoff parameters for the $\ell_2$ and the perceptual losses, respectively.

\textbf{Adversarial Loss}.
Following~\cite{li2017generative,iizuka2017globally}, both global and local adversarial losses are deployed to further improve the perceptual quality of the restoration result.
Let $p_{data}({I})$ be the distribution of ground-truth image, $p_{d}({I^d})$ be the distribution of degraded observation.
Using the global adversarial loss~\cite{Goodfellow2014Generative} as an example, the adversarial loss can be formulated as,
\begin{equation}\label{eqn:globalgan}
\ell_{a,g} \!=\! \min_{\Theta} \max_{D}  \mathbb{E}_{I \sim p_{data}({I})} [\log D({I})] + \mathbb{E}_{I^d \sim p_{d}({I^d})} [\log ( 1 \!-\! D(\mathcal{F}(I^{d}, I^{g}; \Theta)) )],
\end{equation}
where $D(I)$ denotes the global discriminator which predicts the possibility that $I$ is from the distribution $p_{data}({I})$. $\mathcal{F}(I^{d}, I^{g}; \Theta)$ denotes the restoration result by our GFRNet with the model parameters $\Theta = (\Theta_w, \Theta_r)$.

Following the conditional GAN~\cite{pix2pix2016}, the discriminator has the same architecture with pix2pix~\cite{pix2pix2016}, and takes the degraded observation, guided image and restoration result as the input.
The network is trained in an adversarial manner, where our GFRNet is updated by minimizing the loss $\ell_{a,g}$ while the discriminator is updated by maximizing $\ell_{a,g}$.
To improve the training stability, we adopt the improved GAN~\cite{SalimansGZCRC16}, and replace the labels 0/1 with the smoothed 0/0.9 to reduce the vulnerability to adversarial examples.
The local adversarial loss $\ell_{a,l}$ adopts the same settings with the global one but its discriminator is defined only on the minimal bounding box enclosing all facial landmarks.
To sum up, the overall adversarial loss is defined as
\begin{equation}\label{eqn:adversarial}
\mathcal{L}_a = \lambda_{a,g} \ell_{a,g} + \lambda_{a,l} \ell_{a,l}.
\end{equation}
where $\lambda_{a,g}$ and $\lambda_{a,l}$ are the tradeoff parameters for the global and local adversarial losses, respectively.
\subsubsection{Losses on Flow Field $\Phi$.}
Although the WarpNet can be end-to-end trained based on the reconstruction and adversarial losses,
it cannot be learned to correctly align $I^w$ with $I$ in terms of pose and expression (see Fig.~\ref{fig:warp}).
In~\cite{ganin2016deepwarp,yeh2016semantic}, the appearance flow network is trained by minimizing the MSE loss between the output and the ground-truth of the warped image.
But for guided face restoration, $I$ generally has different illumination with $I^g$, and cannot serve as the ground-truth of the warped image.
To circumvent this issue, we present the landmark loss as well as the TV regularization to facilitate the learning of WarpNet.

\textbf{Landmark loss}.
Using the face alignment method TCDCN~\cite{TCDCN}, we detect the 68 landmarks $\{( {{x_j^{I^g}},{y_j^{I^g}}})|_{j=1}^{68}\}$ for $I^g$ and $\{( {{x_j^{I}},{y_j^{I}}})|_{j=1}^{68}\}$ for $I$.
In order to align $I^w$ and $I$, it is natural to require that the landmarks of $I^w$ are close to those of $I$, i.e., $\Phi^x({{x_j^{I}},{y_j^{I}}}) \approx {x_j^{I^g}}$ and $\Phi^y({{x_j^{I}},{y_j^{I}}}) \approx {y_j^{I^g}}$.
Thus, the landmark loss is defined as
\begin{equation}\label{Eqn1}
\ell_{lm} \!=\! \sum_{i} (\Phi_x({x_i^{I}},{y_i^{I}}) - x_i^{I^g} )^2 + (\Phi_y({x_i^{I}},{y_i^{I}}) - y_i^{I^g} )^2.
\end{equation}
In our implementation, all the coordinates (including $x$, $y$, $\Phi_x$ and $\Phi_y$) are normalized to the range $[-1,1]$.

\textbf{TV regularization}.
The landmark loss can only be imposed on the locations of the 68 landmarks.
For better learning WarpNet, we further take the TV regularization into account to require that the flow field should be spatially smooth.
Given the 2D dense flow field $(f_x, f_y)$, the TV regularizer is defined as
\begin{equation}\label{eqn:tv}
\ell_{TV} = \|\nabla_x \Phi_x\|^2 + \|\nabla_y \Phi_x \|^2 + \|\nabla_x \Phi_y \|^2 + \|\nabla_y \Phi_y \|^2,
\end{equation}
where $\nabla_x$ ($\nabla_y$) denotes the gradient operator along the $x$ ($y$) coordinate.

Combining landmark loss with TV regularizer, we define the flow loss as
\begin{equation}\label{eqn:tv}
\mathcal{L}_{flow} = \lambda_{lm} \ell_{lm} + \lambda_{TV} \ell_{TV},
\end{equation}
where $\lambda_{lm}$ and $\lambda_{TV}$ denote the tradeoff parameters for landmark loss  and TV regularizer, respectively.
\subsubsection{Overall Objective.}
Finally, we combine the reconstruction loss, adversarial loss, and flow loss to give the overall objective,
\begin{equation}\label{eqn:objective}
\mathcal{L} = \mathcal{L}_{r} + \mathcal{L}_{a} + \mathcal{L}_{flow}.
\end{equation}
\section{Experimental Results}\label{section4}
Extensive experiments are conducted to assess our GFRNet for guided blind face restoration.
Peak Signal-to-Noise Ratio~(PSNR) and structural similarity~(SSIM) indices are adopted for quantitative evaluation with the related state-of-the-arts (including image super-resolution, deblurring, {denoising, compression artifact removal} and face hallucination).
As for qualitative evaluation, we illustrate the results by our GFRNet and the competing methods.
Results on real low quality images are also given to evaluate the generalization ability of our GFRNet.
More results and testing code are available at: \url{https://github.com/csxmli2016/GFRNet}.
\subsection{Dataset}
We adopt the CASIA-WebFace~\cite{Webface} and VggFace2~\cite{Vggface2} datasets to constitute our training and test sets.
The WebFace contains 10,575 identities and each has about 46 images with the size $256 \times 256$.
The VggFace2 contains 9,131 identities~(8,631 for training and 500 for testing) and each has an average of 362 images with different sizes.
The images in the two datasets are collected in the wild and cover a large range of pose, age, illumination and expression.
For each identity, at most five high quality images are selected, in which a frontal image with eyes open is chosen as the guided image and the others are used as the ground-truth to generate degraded observations.
By this way, we build our training set of 20,273 pairs of ground-truth and guided images from the VggFace2 training set.
Our test set includes two subsets: (i) 1,005 pairs from the VggFace2 test set, and (ii) 1,455 pairs from WebFace.
The images whose identities have appeared in our training set are excluded from the test set.
Furthermore, low quality images are also excluded in training and testing, which include: (i) low-resolution images, (ii) images with large occlusion, (iii) cartoon images, and (iv) images with obvious artifacts.
The face region of each image in VGGFace2 is cropped and resized to $256 \times 256$ based on the bounding box detected by MTCNN~\cite{MTCNN}.
All training and test images are not aligned to keep their original pose and expression.
Facial landmarks of the ground-truth and guided images are detected by TCDCN~\cite{TCDCN} and are only used in training.

\subsection{Training Details and Parameter Setting}
Our model is trained using the Adam algorithm~\cite{kingma2014adam} with the learning rate of $2 \times {10^{ - 4}}$, $2 \times {10^{ - 5}}$, $2 \times {10^{ - 6}}$  and ${\beta _1} = 0.5$.
In each learning rate, the model is trained until the reconstruction loss becomes non-decreasing.
Then a smaller learning rate is adopted to further fine-tune the model.
The tradeoff parameters are set as $\lambda_{r,0} = 100$, $\lambda_{r,l} = 0.001$, $\lambda_{a,g} = 1$, $\lambda_{a,l} = 0.5$, $\lambda_{lm} = 10$, and $\lambda_{TV} = 1$.
We first pre-train the WarpNet for 5 epochs by minimizing the flow loss $\mathcal{L}_{flow}$, and then both WarpNet and RecNet are end-to-end trained by using the objective $\mathcal{L}$.
The batch size is 1 and the training is stopped after 100 epochs.
Data augmentation such as flipping is also adopted during training.
\subsection{Results on Synthetic Images}
\begin{table*}[t]
	{
		\scriptsize 
		\caption{Quantitative results on two test subsets. Numbers in the parentheses indicate SSIM and the remaining represents PSNR (dB). The best results are highlighted in {\bf \color{red}red} and second best ones except our GFRNet variants are highlighted in {\color{blue}blue}.
		}
		\begin{center}
			\label{table::Quantization}
			\begin{tabular}{ c| l|c c c  c|c c c c}
				\hline
				\multicolumn{2}{c|}{\multirow{2}{*}{\makecell[c]{Methods}}} & \multicolumn{4}{c}{VggFace2~\cite{Vggface2}} &  \multicolumn{4}{|c}{WebFace~\cite{Webface}}\\
				\cline{3-10}
				\multicolumn{2}{c|}{}&\multicolumn{2}{c}{$4\times$} &  \multicolumn{2}{c}{$8\times$} &\multicolumn{2}{|c}{$4\times$} &  \multicolumn{2}{c}{$8\times$}\\
				\hline
				\multirow{4}{*}{SR}&SRCNN~\cite{dong2014learning}& 24.57 & (.842) & 22.30 & (.802)& 26.11 & (.872) & 23.50 & (.842) \\
				&VDSR~\cite{kim2016accurate}& 25.36 & (.858) & 22.50 & (.807)& 26.60 & (.884) &  23.65 & (.847)\\
				&SRGAN~\cite{Ledig2017CVPR}& 25.85 & (.911) & 23.01 & (.874)& 27.65 & (.941) &  24.49 & (.913)\\
				
				&MSRGAN& 26.55 & (.906) &  {\color{blue}23.45} & (.862) & 28.10 & (.934) & 24.92 & (.908)\\
				\hline
				\multirow{4}{*}{Deblurring}&DCP~\cite{pan2016blind}& 24.42 & (.894) & 21.54 &(.848) & 24.97 & (.895) & 23.05 &(.887)\\
				&DeepDeblur~\cite{Nah2017CVPR}& 26.31 & (.917) & 22.97 &(.873) & 28.13 & (.934) & 24.63 &(.910)\\
				&DeblurGAN~\cite{DeblurGAN}& 24.65 & (.889) & 22.06 &(.846) & 24.63 & (.910) & 23.38 &(.896)\\
				&MDeblurGAN& 25.32 & (.918) & 22.46 &(.867) & {\color{blue}29.41} & ({\color{blue}.952}) & 23.49 &(.900)\\
				\hline
				\multirow{3}{*}{Denoising}&DnCNN~\cite{zhang2017beyond}& 26.73 & (.920) & 23.29 &(.877) & 28.35 & (.933) & 24.75 &(.912)\\
				&MemNet~\cite{MemNet}& 26.85 & (.923) & 23.31 &(.877) & 28.57 & (.934) & 24.77 &(.909)\\
                &MDnCNN& {\color{blue}27.05} & ({\color{blue}.925}) & 23.33 &({\color{blue}.879}) & 29.40 & (.942) & 24.84 &(.912)\\
				\hline
				\multirow{2}{*}{AR}&ARCNN~\cite{Dong2015ICCV}& 22.05 & (.863) & 20.84 &(.827) & 23.39 & (.876) & 20.47 &(.858)\\
				&MARCNN& 25.43 & (.923) & 23.16 &(.876) & 28.40 & (.938) & {\color{blue}25.15} &({\color{blue}.914})\\
				\hline
				\multirow{3}{*}{Non-blind FH}&CBN~\cite{zhu2016deep}& 24.52 & (.867)& 21.84 & (.817)& 25.43 & (.899) & 23.10 & (.852)\\
				&WaveletSRNet~\cite{huang2017wavelet}& 25.66 & (.909) & 20.87 &(.831) & 27.10 & (.937) & 21.63 &(.869)\\
				&TDAE~\cite{yu2017hallucinating}& - & (-) & 20.19 &(.729) & - & (-) & 20.24 &(.741)\\
				\hline
				\multirow{2}{*}{Blind FH}&SCGAN~\cite{xu2017learning}& 25.16 & (.905) & - & -& 26.37 & (.923) & - & -\\
				&MCGAN~\cite{xu2017learning}& 25.26 & (.912) & - & -& 26.35 & (.931) & - & - \\
				\hline
				\multirow{7}{*}{Ours}&Ours($-WG$)& 25.97 & (.915) & 22.91 & (.838) & 28.73 & (.928) & 24.76 & (.884)\\
				&Ours($-WG2$)& 27.20 & (.932) & 23.22 & (.863) & 29.45 & (.945) & 25.93 & (.914)\\
				&Ours($-W$) & 26.03 & (.923) & 23.29 & (.843) & 29.66 & (.934) & 25.20 & (.897)\\
				&Ours($-W2$) & 27.25 & (.933) & 23.24 & (.864) & 29.73 & (.948)& 25.95 & (.917) \\
				&Ours($-F$)& 26.61 & (.927) & 23.17 & (.863) & 31.43 & (.920) & 26.00 & (.922)\\
				&Ours($R$)& 27.90 & (.943) & 24.05 & (.890) & 31.46 & (.962) & 26.88 & (.922)\\
				&{\bf Ours($Full$)}& {\bf \color{red}28.55} & ({\bf \color{red}.947}) & {\bf \color{red}24.10} & ({\bf \color{red}.898})& {\bf \color{red}32.31} & ({\bf \color{red}.973}) & {\bf \color{red}27.21} & ({\bf \color{red}.935})\\
				\hline
			\end{tabular}
		\end{center}
	}
\end{table*}
Table~\ref{table::Quantization} lists the PSNR and SSIM results on the two test subsets, where our GFRNet achieves significant performance gains over all the competing methods.
Using the $4\times$ SR on WebFace as an example, in terms of PSNR, our GFRNet outperforms the {SR} and blind deblurring methods by more than 4 dB, the denoising methods by more than 3.5 dB, the {compression artifact removal (AR)} methods by more than 8 dB, the non-blind and blind {FH} methods by more than 5 dB.
To the best of our knowledge, guided blind face restoration remains an uninvestigated issue in literature.
Thus, we compare our GFRNet with several relevant state-of-the-arts, including three non-blind image super-resolution (SR) methods (SRCNN~\cite{dong2014learning}, VDSR~\cite{kim2016accurate}, SRGAN~\cite{Ledig2017CVPR}), three blind deblurring methods (DCP~\cite{pan2016blind}, DeepDeblur~\cite{Nah2017CVPR}, DeblurGAN~\cite{DeblurGAN}), two denoising methods (DnCNN~\cite{zhang2017beyond}, MemNet~\cite{MemNet}), one compression artifact removal method (ARCNN~\cite{Dong2015ICCV}), three non-blind face hallucination (FH) methods (CBN~\cite{zhu2016deep}, WaveletSRNet~\cite{huang2017wavelet}, TDAE~\cite{yu2017hallucinating}),
and two blind FH methods (SCGAN~\cite{xu2017learning}, MCGAN~\cite{xu2017learning}).
To keep consistent with the SR and FH methods, only two scale factors, i.e., 4 and 8, are considered for the test images.
As for non-SR methods, we take the bicubic upsampling result as the input to the model.
SRCNN~\cite{dong2014learning} and VDSR~\cite{kim2016accurate} are only trained to perform $2\times$, $3\times$ and $4\times$ SR.
To handle $8\times$ SR, we adopt the strategy in~\cite{tuzel2016global} by applying the $2\times$ model to the result produced by the $4\times$ model.
For SCGAN~\cite{xu2017learning} and MCGAN~\cite{xu2017learning}, only the $4\times$ models are available. For TDAE~\cite{yu2017hallucinating}, only the $8\times$ model is available.

\subsubsection{Quantitative evaluation.}
It is worth noting that the promising performance of our GFRNet cannot be solely attributed to the use of our training data and the simple incorporation of guided image.
To illustrate this point, we retrain four competing methods (i.e., SRGAN, DeblurGAN, DnCNN, and ARCNN) by using our training data and taking both degraded observation and guided image as input.
For the sake of distinction, the retrained models are represented as MSRGAN{, MDeblurGAN, MDnCNN, MARCNN}.
From Table~\ref{table::Quantization}, the retrained models do achieve better PSNR and SSIM results than the original ones, but still perform inferior to our GFRNet with a large margin, especially on WebFace.
Therefore, the performance gains over the retrained models should be explained by the network architecture and model objective of our GFRNet.

\subsubsection{Qualitative evaluation.}
\begin{figure}[t]
\centering
\subfigure[\scriptsize Input]{
  \begin{minipage}[b]{.19\columnwidth}
    \includegraphics[width=0.99\textwidth]{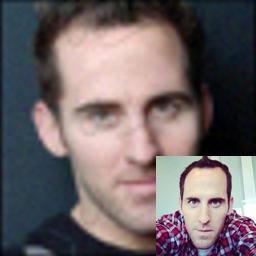}
  \end{minipage}
}
\vspace{-1.5ex}
\hspace{-3ex}
\subfigure[\scriptsize SRCNN\cite{dong2014learning}]{
  \begin{minipage}[b]{.19\columnwidth}
    \includegraphics[width=0.99\textwidth]{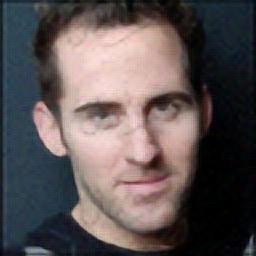}
  \end{minipage}
}
\hspace{-3ex}
\subfigure[\scriptsize VDSR\cite{kim2016accurate}]{
  \begin{minipage}[b]{.19\columnwidth}
    \includegraphics[width=0.99\textwidth]{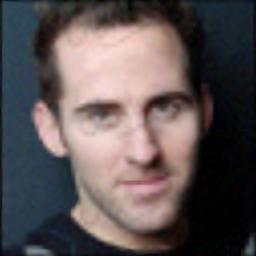}
  \end{minipage}
}
\hspace{-3ex}
\subfigure[\scriptsize SRGAN\cite{Ledig2017CVPR}]{
  \begin{minipage}[b]{.19\columnwidth}
    \includegraphics[width=0.99\textwidth]{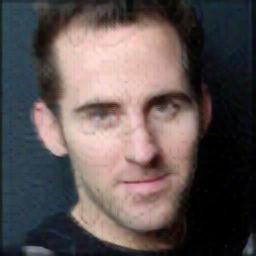}
  \end{minipage}
}
\hspace{-3ex}
\subfigure[\scriptsize MSRGAN]{
  \begin{minipage}[b]{.19\columnwidth}
    \includegraphics[width=0.99\textwidth]{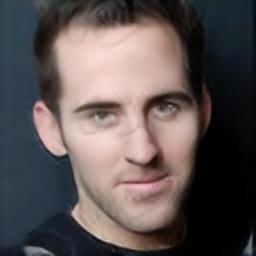}
  \end{minipage}
}
\hspace{-3ex}
\subfigure[\scriptsize DCP\cite{pan2016blind}]{
  \begin{minipage}[b]{.19\columnwidth}
    \includegraphics[width=0.99\textwidth]{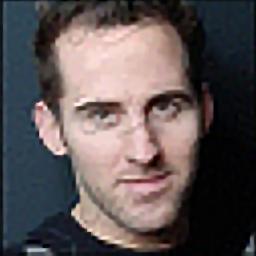}
  \end{minipage}
}
\vspace{-1.5ex}
\hspace{-3ex}
\subfigure[{\fontsize{6}{5}\selectfont DeepDeblur\cite{Nah2017CVPR}}]{
  \begin{minipage}[b]{.19\columnwidth}
    \includegraphics[width=0.99\textwidth]{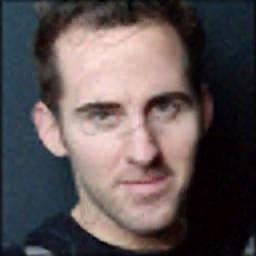}
  \end{minipage}
}
\hspace{-3ex}
\subfigure[{\fontsize{6}{5}\selectfont DeblurGAN\cite{DeblurGAN}}]{
  \begin{minipage}[b]{.19\columnwidth}
    \includegraphics[width=0.99\textwidth]{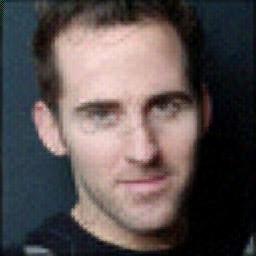}
  \end{minipage}
}
\hspace{-3ex}
\subfigure[\scriptsize MDeblurGAN]{
  \begin{minipage}[b]{.19\columnwidth}
    \includegraphics[width=0.99\textwidth]{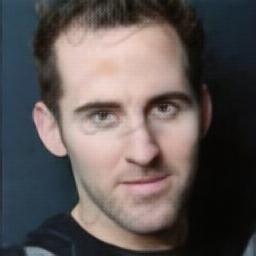}
  \end{minipage}
}
\hspace{-3ex}
\subfigure[\scriptsize MemNet\cite{MemNet}]{
  \begin{minipage}[b]{.19\columnwidth}
    \includegraphics[width=0.99\textwidth]{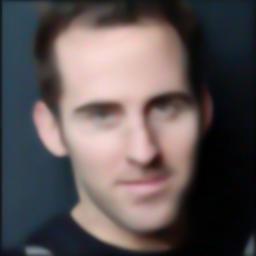}
  \end{minipage}
}
\hspace{-3ex}
\subfigure[\scriptsize DnCNN\cite{zhang2017beyond}]{
  \begin{minipage}[b]{.19\columnwidth}
    \includegraphics[width=0.99\textwidth]{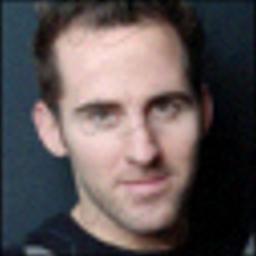}
  \end{minipage}
}
\vspace{-1.5ex}
\hspace{-3ex}
\subfigure[\scriptsize MDnCNN]{
  \begin{minipage}[b]{.19\columnwidth}
    \includegraphics[width=0.99\textwidth]{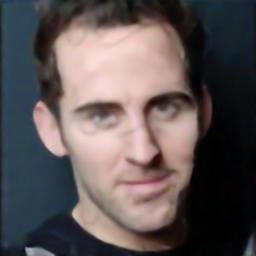}
  \end{minipage}
}
\hspace{-3ex}
\subfigure[\scriptsize ARCNN\cite{Dong2015ICCV}]{
  \begin{minipage}[b]{.19\columnwidth}
    \includegraphics[width=0.99\textwidth]{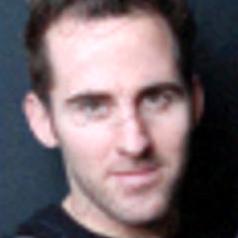}
  \end{minipage}
}
\hspace{-3ex}
\subfigure[\scriptsize MARCNN]{
  \begin{minipage}[b]{.19\columnwidth}
    \includegraphics[width=0.99\textwidth]{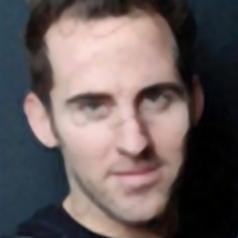}
  \end{minipage}
}
\hspace{-3ex}
\subfigure[\scriptsize CBN\cite{zhu2016deep}]{
  \begin{minipage}[b]{.19\columnwidth}
    \includegraphics[width=0.99\textwidth]{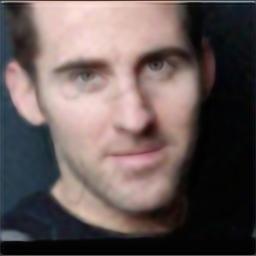}
  \end{minipage}
}
\hspace{-3ex}
\subfigure[{\fontsize{6}{5}\selectfont WaveletSRNet\cite{huang2017wavelet}}]{
  \begin{minipage}[b]{.19\columnwidth}
    \includegraphics[width=0.99\textwidth]{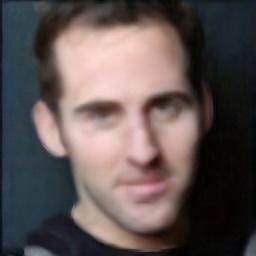}
  \end{minipage}
}
\hspace{-3ex}
\subfigure[\scriptsize SCGAN\cite{xu2017learning}]{
  \begin{minipage}[b]{.19\columnwidth}
    \includegraphics[width=0.99\textwidth]{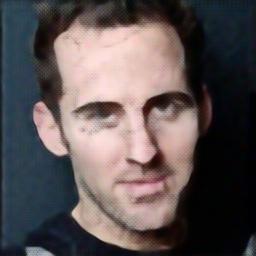}
  \end{minipage}
}
\hspace{-3ex}
\subfigure[\scriptsize MCGAN\cite{xu2017learning}]{
  \begin{minipage}[b]{.19\columnwidth}
    \includegraphics[width=0.99\textwidth]{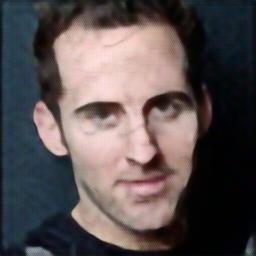}
  \end{minipage}
}
\hspace{-3ex}
\subfigure[\scriptsize Ours]{
  \begin{minipage}[b]{.19\columnwidth}
    \includegraphics[width=0.99\textwidth]{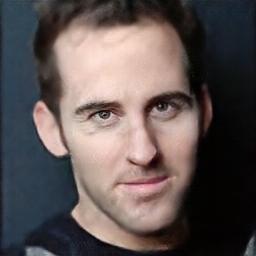}
  \end{minipage}
}
\hspace{-3ex}
\subfigure[\scriptsize Ground-truth]{
  \begin{minipage}[b]{.19\columnwidth}
    \includegraphics[width=0.99\textwidth]{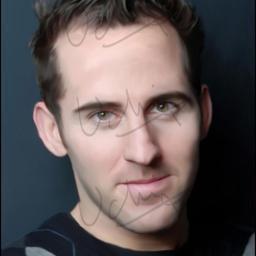}
  \end{minipage}
}
\vspace{-1ex}
\caption{The $4 \times$ SR results compared with all the competing methods.}
\label{fig:x4all}
\end{figure}
\begin{figure}[htb]
\centering
\subfigure[\scriptsize Input]{
  \begin{minipage}[b]{.19\columnwidth}
    \includegraphics[width=0.99\textwidth]{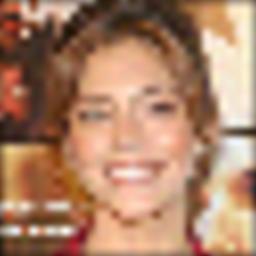}
  \end{minipage}
}
\vspace{-1.5ex}
\hspace{-3ex}
\subfigure[\scriptsize Guided image]{
  \begin{minipage}[b]{.19\columnwidth}
    \includegraphics[width=0.99\textwidth]{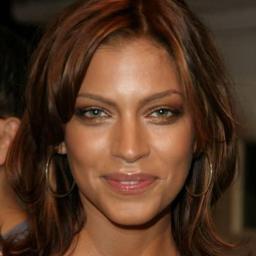}
  \end{minipage}
}
\hspace{-3ex}
\subfigure[\scriptsize SRCNN\cite{dong2014learning}]{
  \begin{minipage}[b]{.19\columnwidth}
    \includegraphics[width=0.99\textwidth]{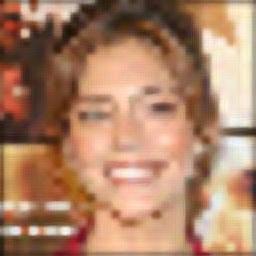}
  \end{minipage}
}
\hspace{-3ex}
\subfigure[\scriptsize VDSR\cite{kim2016accurate}]{
  \begin{minipage}[b]{.19\columnwidth}
    \includegraphics[width=0.99\textwidth]{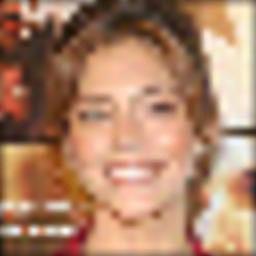}
  \end{minipage}
}
\hspace{-3ex}
\subfigure[\scriptsize SRGAN\cite{Ledig2017CVPR}]{
  \begin{minipage}[b]{.19\columnwidth}
    \includegraphics[width=0.99\textwidth]{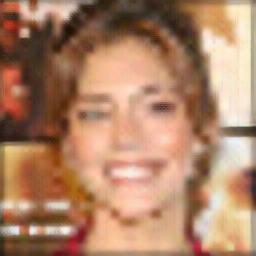}
  \end{minipage}
}
\hspace{-3ex}
\subfigure[\scriptsize MSRGAN]{
  \begin{minipage}[b]{.19\columnwidth}
    \includegraphics[width=0.99\textwidth]{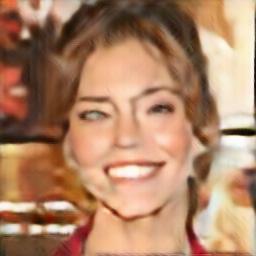}
  \end{minipage}
}
\vspace{-1.5ex}
\hspace{-3ex}
\subfigure[\scriptsize DCP\cite{pan2016blind}]{
  \begin{minipage}[b]{.19\columnwidth}
    \includegraphics[width=0.99\textwidth]{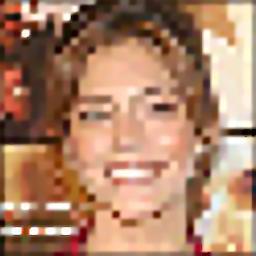}
  \end{minipage}
}
\hspace{-3ex}
\subfigure[{\fontsize{6}{5}\selectfont DeepDeblur\cite{Nah2017CVPR}}]{
  \begin{minipage}[b]{.19\columnwidth}
    \includegraphics[width=0.99\textwidth]{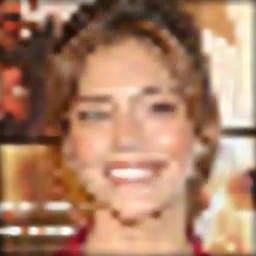}
  \end{minipage}
}
\hspace{-3ex}
\subfigure[{\fontsize{6}{5}\selectfont DeblurGAN\cite{DeblurGAN}}]{
  \begin{minipage}[b]{.19\columnwidth}
    \includegraphics[width=0.99\textwidth]{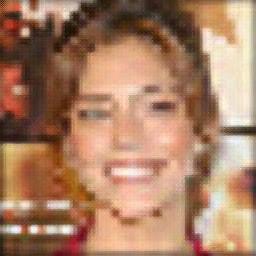}
  \end{minipage}
}
\hspace{-3ex}
\subfigure[\scriptsize MDeblurGAN]{
  \begin{minipage}[b]{.19\columnwidth}
    \includegraphics[width=0.99\textwidth]{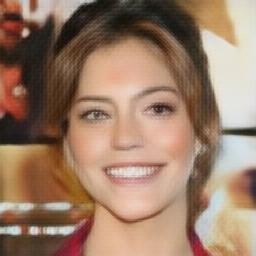}
  \end{minipage}
}
\hspace{-3ex}
\subfigure[\scriptsize MemNet\cite{MemNet}]{
  \begin{minipage}[b]{.19\columnwidth}
    \includegraphics[width=0.99\textwidth]{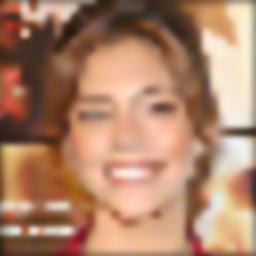}
  \end{minipage}
}
\vspace{-1.5ex}
\hspace{-3ex}
\subfigure[\scriptsize DnCNN\cite{zhang2017beyond}]{
  \begin{minipage}[b]{.19\columnwidth}
    \includegraphics[width=0.99\textwidth]{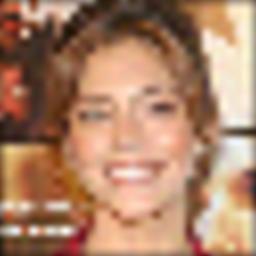}
  \end{minipage}
}
\hspace{-3ex}
\subfigure[\scriptsize MDnCNN]{
  \begin{minipage}[b]{.19\columnwidth}
    \includegraphics[width=0.99\textwidth]{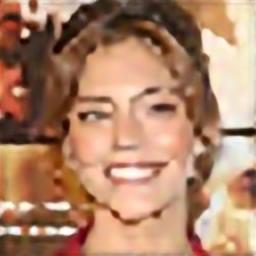}
  \end{minipage}
}
\hspace{-3ex}
\subfigure[\scriptsize ARCNN\cite{Dong2015ICCV}]{
  \begin{minipage}[b]{.19\columnwidth}
    \includegraphics[width=0.99\textwidth]{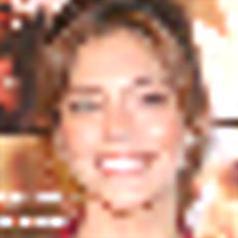}
  \end{minipage}
}
\hspace{-3ex}
\subfigure[\scriptsize MARCNN]{
  \begin{minipage}[b]{.19\columnwidth}
    \includegraphics[width=0.99\textwidth]{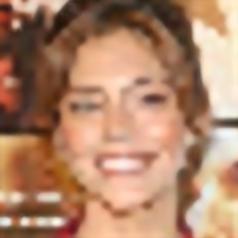}
  \end{minipage}
}
\hspace{-3ex}
\subfigure[\scriptsize CBN\cite{zhu2016deep}]{
  \begin{minipage}[b]{.19\columnwidth}
    \includegraphics[width=0.99\textwidth]{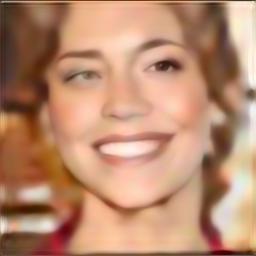}
  \end{minipage}
}
\hspace{-3ex}
\subfigure[{\fontsize{6}{5}\selectfont WaveletSRNet\cite{huang2017wavelet}}]{
  \begin{minipage}[b]{.19\columnwidth}
    \includegraphics[width=0.99\textwidth]{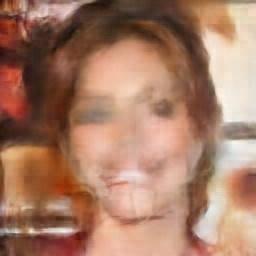}
  \end{minipage}
}
\hspace{-3ex}
\subfigure[\scriptsize TDAE\cite{yu2017hallucinating}]{
  \begin{minipage}[b]{.19\columnwidth}
    \includegraphics[width=0.99\textwidth]{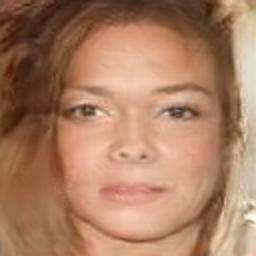}
  \end{minipage}
}
\hspace{-3ex}
\subfigure[\scriptsize Ours]{
  \begin{minipage}[b]{.19\columnwidth}
    \includegraphics[width=0.99\textwidth]{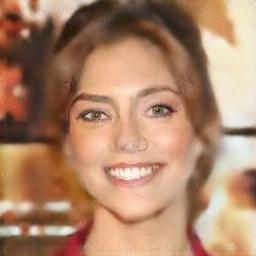}
  \end{minipage}
}
\hspace{-3ex}
\subfigure[\scriptsize Ground-truth]{
  \begin{minipage}[b]{.19\columnwidth}
    \includegraphics[width=0.99\textwidth]{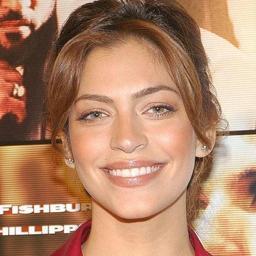}
  \end{minipage}
}
\vspace{-1ex}
\caption{The $8 \times$ SR results compared with all the competing methods.}
\label{fig:x8all}
\end{figure}
%
\begin{figure}[t]
\centering
\subfigure[]{
  \begin{minipage}[b]{.16\columnwidth}
    \includegraphics[width=0.99\textwidth]{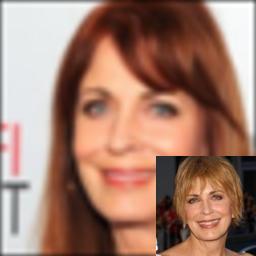}\\
    \includegraphics[width=0.99\textwidth]{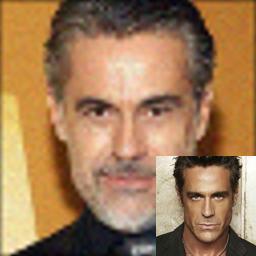}\\
    \includegraphics[width=0.99\textwidth]{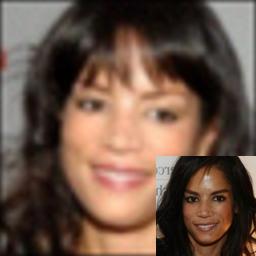}\\
    \includegraphics[width=0.99\textwidth]{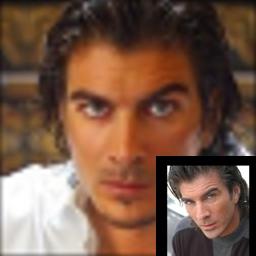}
  \end{minipage}
}
\hspace{-3ex}
\subfigure[]{
  \begin{minipage}[b]{.16\columnwidth}
    \includegraphics[width=0.99\textwidth]{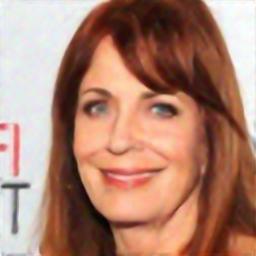}\\
    \includegraphics[width=0.99\textwidth]{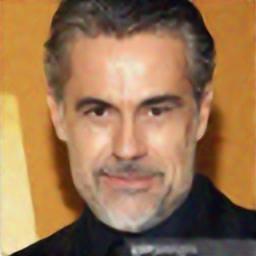}\\
    \includegraphics[width=0.99\textwidth]{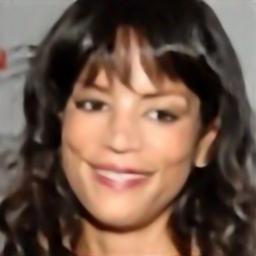}\\
    \includegraphics[width=0.99\textwidth]{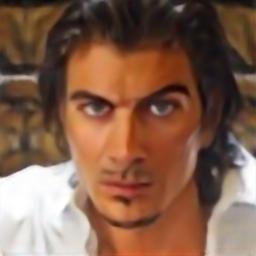}
  \end{minipage}
}
\hspace{-3ex}
\subfigure[]{
  \begin{minipage}[b]{.16\columnwidth}
    \includegraphics[width=0.99\textwidth]{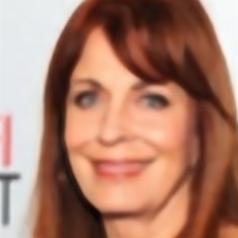}\\
    \includegraphics[width=0.99\textwidth]{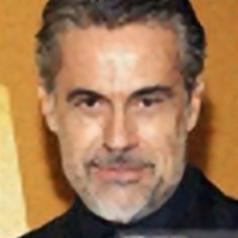}\\
    \includegraphics[width=0.99\textwidth]{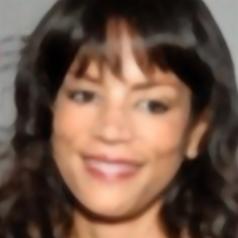}\\
    \includegraphics[width=0.99\textwidth]{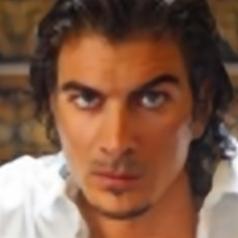}
  \end{minipage}
}
\hspace{-3ex}
\subfigure[]{
  \begin{minipage}[b]{.16\columnwidth}
    \includegraphics[width=0.99\textwidth]{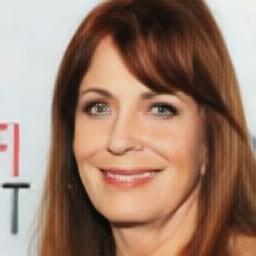}\\
    \includegraphics[width=0.99\textwidth]{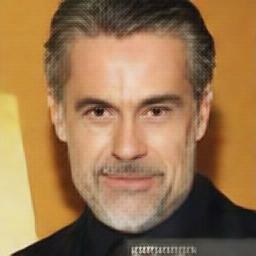}\\
    \includegraphics[width=0.99\textwidth]{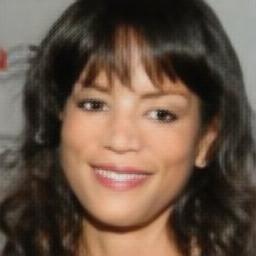}\\
    \includegraphics[width=0.99\textwidth]{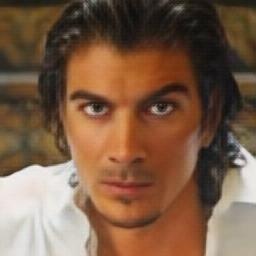}
  \end{minipage}
}
\hspace{-3ex}
\subfigure[]{
  \begin{minipage}[b]{.16\columnwidth}
    \includegraphics[width=0.99\textwidth]{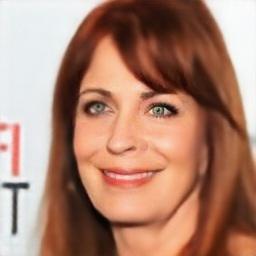}\\
    \includegraphics[width=0.99\textwidth]{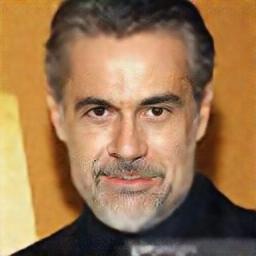}\\
    \includegraphics[width=0.99\textwidth]{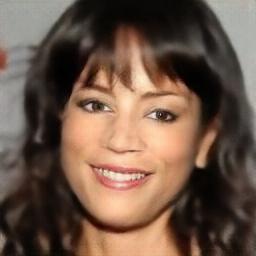}\\
    \includegraphics[width=0.99\textwidth]{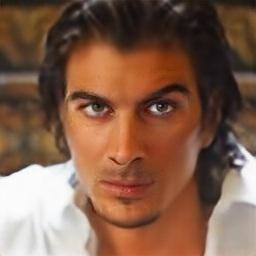}

  \end{minipage}
}
\hspace{-3ex}
\subfigure[]{
  \begin{minipage}[b]{.16\columnwidth}
    \includegraphics[width=0.99\textwidth]{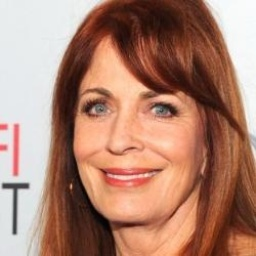}\\
    \includegraphics[width=0.99\textwidth]{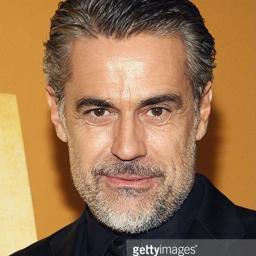}\\
    \includegraphics[width=0.99\textwidth]{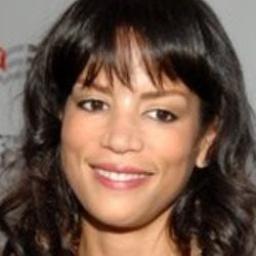}\\
    \includegraphics[width=0.99\textwidth]{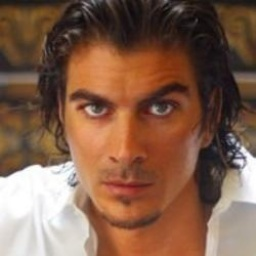}

  \end{minipage}
}
  \caption{The 4$\times$ SR results: (a)~synthetic low quality image (Close-up in right bottom is the guided image), (b)~MDnCNN~\cite{zhang2017beyond}, (c)~MARCNN~\cite{Dong2015ICCV}, (d)~MDeblurGAN~\cite{DeblurGAN}, (e)~Ours, and (f)~ground-truth.}
  \label{fig:x4}
\end{figure}
\begin{figure}[htb]
\centering
\subfigure[]{
  \begin{minipage}[b]{.16\columnwidth}
    \includegraphics[width=0.99\textwidth]{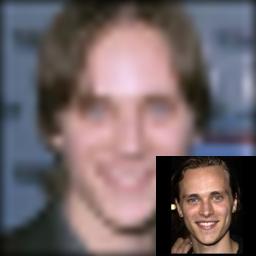}\\
    \includegraphics[width=0.99\textwidth]{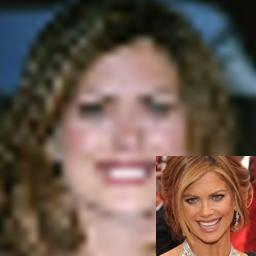}\\
    \includegraphics[width=0.99\textwidth]{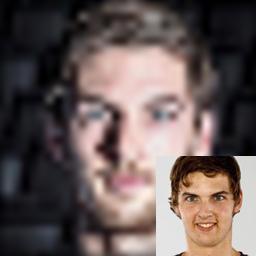}\\
    \includegraphics[width=0.99\textwidth]{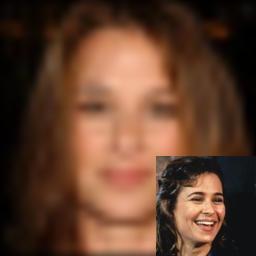}
  \end{minipage}
}
\hspace{-3ex}
\subfigure[]{
  \begin{minipage}[b]{.16\columnwidth}
    \includegraphics[width=0.99\textwidth]{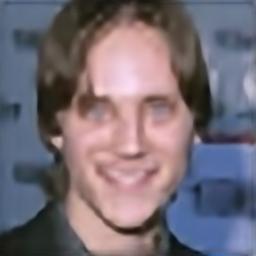}\\
    \includegraphics[width=0.99\textwidth]{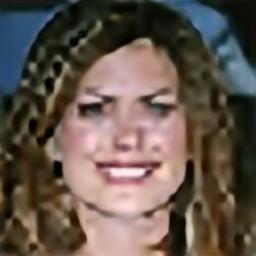}\\
    \includegraphics[width=0.99\textwidth]{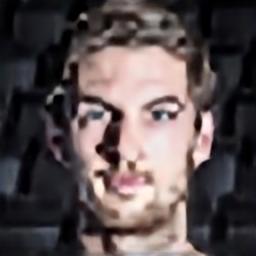}\\
    \includegraphics[width=0.99\textwidth]{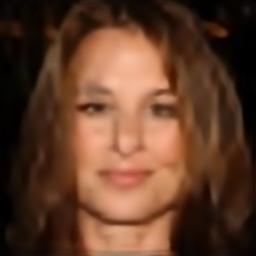}
  \end{minipage}
}
\hspace{-3ex}
\subfigure[]{
  \begin{minipage}[b]{.16\columnwidth}
    \includegraphics[width=0.99\textwidth]{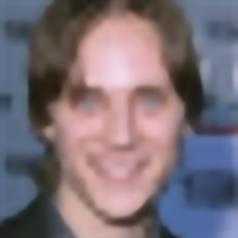}\\
    \includegraphics[width=0.99\textwidth]{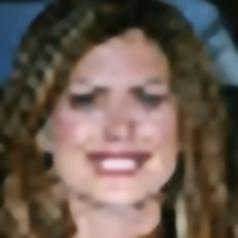}\\
    \includegraphics[width=0.99\textwidth]{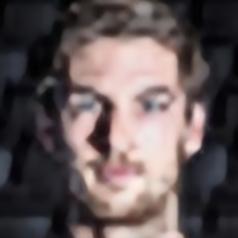}\\
    \includegraphics[width=0.99\textwidth]{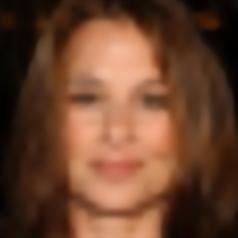}
  \end{minipage}
}
\hspace{-3ex}
\subfigure[]{
  \begin{minipage}[b]{.16\columnwidth}
    \includegraphics[width=0.99\textwidth]{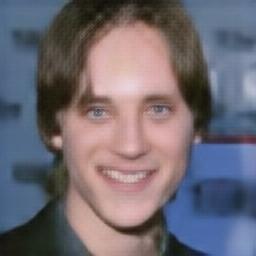}\\
    \includegraphics[width=0.99\textwidth]{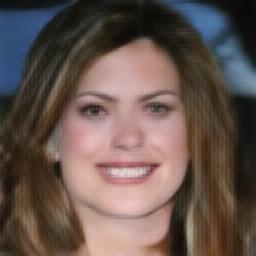}\\
    \includegraphics[width=0.99\textwidth]{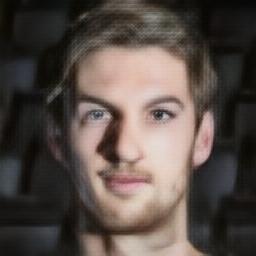}\\
    \includegraphics[width=0.99\textwidth]{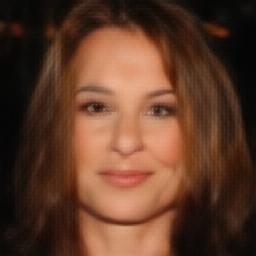}
  \end{minipage}
}
\hspace{-3ex}
\subfigure[]{
  \begin{minipage}[b]{.16\columnwidth}
    \includegraphics[width=0.99\textwidth]{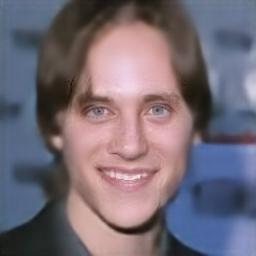}\\
    \includegraphics[width=0.99\textwidth]{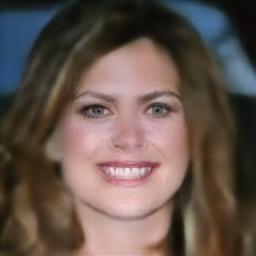}\\
    \includegraphics[width=0.99\textwidth]{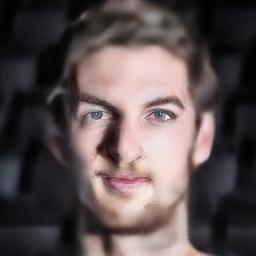}\\
    \includegraphics[width=0.99\textwidth]{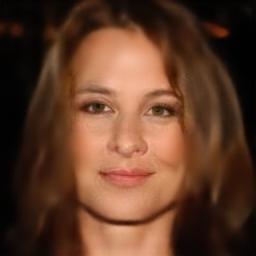}
  \end{minipage}
}
\hspace{-3ex}
\subfigure[]{
  \begin{minipage}[b]{.16\columnwidth}
    \includegraphics[width=0.99\textwidth]{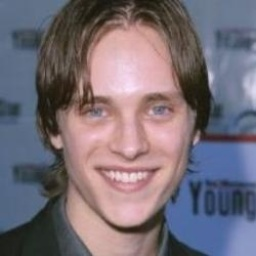}\\
    \includegraphics[width=0.99\textwidth]{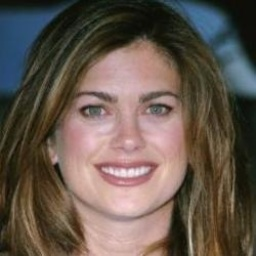}\\
    \includegraphics[width=0.99\textwidth]{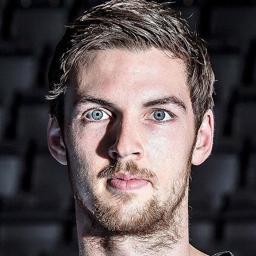}\\
    \includegraphics[width=0.99\textwidth]{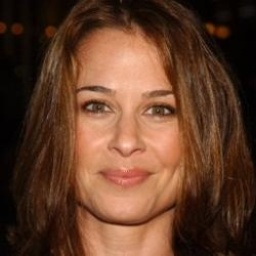}
  \end{minipage}
}
\vspace{-2ex}
  \caption{The 8$\times$ SR results: (a)~synthetic low quality image (Close-up in right bottom is the guided image), (b)~MDnCNN~\cite{zhang2017beyond}, (c)~MARCNN~\cite{Dong2015ICCV}, (d)~MDeblurGAN~\cite{DeblurGAN}, (e)~Ours, and (f)~ground-truth.}
  \label{fig:x8}
\end{figure}
In Figs.~\ref{fig:x4all} and~\ref{fig:x8all}, we compare results of all the competing methods in 4$\times$ SR and 8$\times$ SR. For better comparison, we select three competing methods with top quantitative performance, and compare their results with those by our GFRNet shown in Figs.~\ref{fig:x4} and~\ref{fig:x8}.
It is obvious that our GFRNet is more effective in restoring fine details while suppressing visual artifacts.
In comparison with the competing methods, the results by GFRNet are visually photo-realistic and can correctly recover more fine and identity-aware details especially in eyes, nose, and mouth regions.

\subsection{Results on Real Low Quality Images}
\begin{figure}[t]
\centering
\subfigure[\scriptsize Input]{
  \begin{minipage}[b]{.19\columnwidth}
    \includegraphics[width=0.99\textwidth]{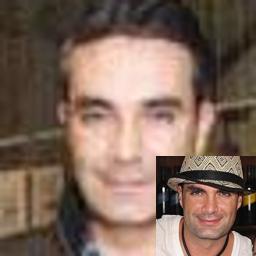}
  \end{minipage}
}
\vspace{-1.5ex}
\hspace{-3ex}
\subfigure[\scriptsize SRCNN\cite{dong2014learning}]{
  \begin{minipage}[b]{.19\columnwidth}
    \includegraphics[width=0.99\textwidth]{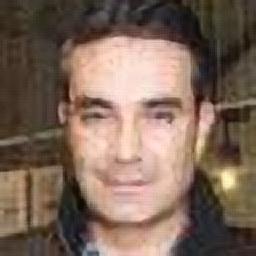}
  \end{minipage}
}
\hspace{-3ex}
\subfigure[\scriptsize VDSR\cite{kim2016accurate}]{
  \begin{minipage}[b]{.19\columnwidth}
    \includegraphics[width=0.99\textwidth]{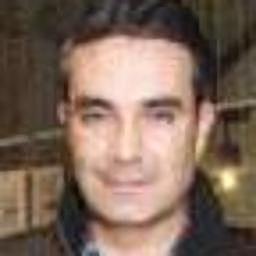}
  \end{minipage}
}
\hspace{-3ex}
\subfigure[\scriptsize SRGAN\cite{Ledig2017CVPR}]{
  \begin{minipage}[b]{.19\columnwidth}
    \includegraphics[width=0.99\textwidth]{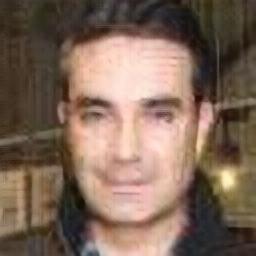}
  \end{minipage}
}
\hspace{-3ex}
\subfigure[\scriptsize MSRGAN]{
  \begin{minipage}[b]{.19\columnwidth}
    \includegraphics[width=0.99\textwidth]{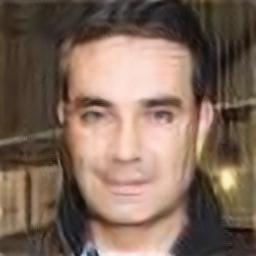}
  \end{minipage}
}
\hspace{-3ex}
\subfigure[\scriptsize DCP\cite{pan2016blind}]{
  \begin{minipage}[b]{.19\columnwidth}
    \includegraphics[width=0.99\textwidth]{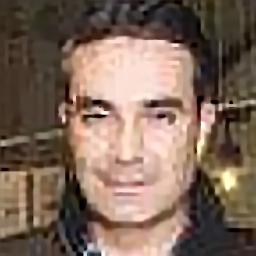}
  \end{minipage}
}
\hspace{-3ex}
\subfigure[{\fontsize{6}{5}\selectfont DeepDeblur\cite{Nah2017CVPR}}]{
  \begin{minipage}[b]{.19\columnwidth}
    \includegraphics[width=0.99\textwidth]{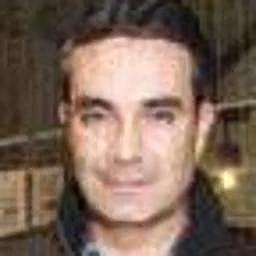}
  \end{minipage}
}
\hspace{-3ex}
\subfigure[{\fontsize{6}{5}\selectfont DeblurGAN\cite{DeblurGAN}}]{
  \begin{minipage}[b]{.19\columnwidth}
    \includegraphics[width=0.99\textwidth]{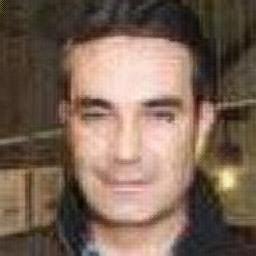}
  \end{minipage}
}
\hspace{-3ex}
\subfigure[\scriptsize MDeblurGAN]{
  \begin{minipage}[b]{.19\columnwidth}
    \includegraphics[width=0.99\textwidth]{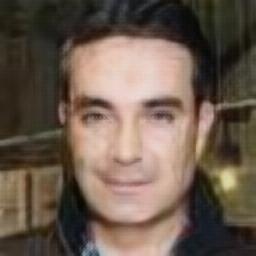}
  \end{minipage}
}
\hspace{-3ex}
\subfigure[\scriptsize MemNet\cite{MemNet}]{
  \begin{minipage}[b]{.19\columnwidth}
    \includegraphics[width=0.99\textwidth]{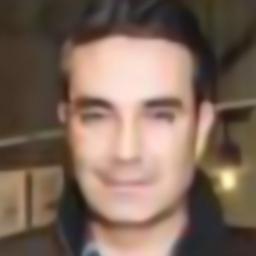}
  \end{minipage}
}
\hspace{-3ex}
\subfigure[\scriptsize DnCNN\cite{zhang2017beyond}]{
  \begin{minipage}[b]{.19\columnwidth}
    \includegraphics[width=0.99\textwidth]{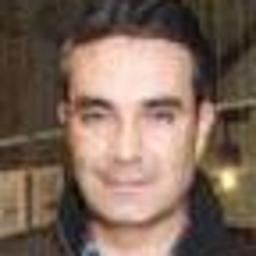}
  \end{minipage}
}
\hspace{-3ex}
\subfigure[\scriptsize MDnCNN]{
  \begin{minipage}[b]{.19\columnwidth}
    \includegraphics[width=0.99\textwidth]{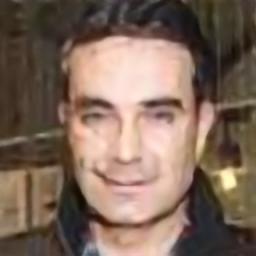}
  \end{minipage}
}
\hspace{-3ex}
\subfigure[\scriptsize ARCNN\cite{Dong2015ICCV}]{
  \begin{minipage}[b]{.19\columnwidth}
    \includegraphics[width=0.99\textwidth]{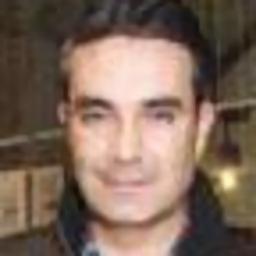}
  \end{minipage}
}
\hspace{-3ex}
\subfigure[\scriptsize MARCNN]{
  \begin{minipage}[b]{.19\columnwidth}
    \includegraphics[width=0.99\textwidth]{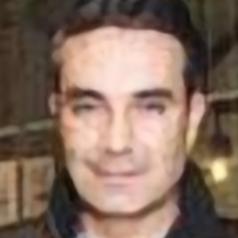}
  \end{minipage}
}
\hspace{-3ex}
\subfigure[\scriptsize CBN\cite{zhu2016deep}]{
  \begin{minipage}[b]{.19\columnwidth}
    \includegraphics[width=0.99\textwidth]{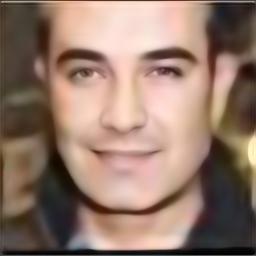}
  \end{minipage}
}
\hspace{-3ex}
\subfigure[{\fontsize{6}{5}\selectfont WaveletSRNet\cite{huang2017wavelet}}]{
  \begin{minipage}[b]{.19\columnwidth}
    \includegraphics[width=0.99\textwidth]{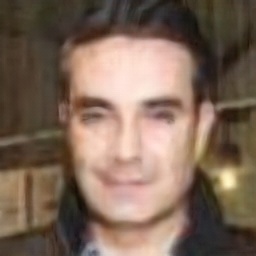}
  \end{minipage}
}
\hspace{-3ex}
\subfigure[\scriptsize TDAE\cite{yu2017hallucinating}]{
  \begin{minipage}[b]{.19\columnwidth}
    \includegraphics[width=0.99\textwidth]{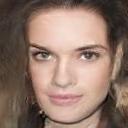}
  \end{minipage}
}
\hspace{-3ex}
\subfigure[\scriptsize SCGAN\cite{xu2017learning}]{
  \begin{minipage}[b]{.19\columnwidth}
    \includegraphics[width=0.99\textwidth]{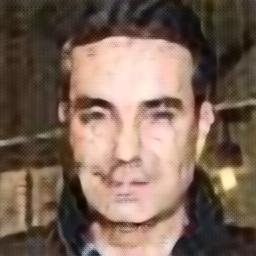}
  \end{minipage}
}
\hspace{-3ex}
\subfigure[\scriptsize MCGAN\cite{xu2017learning}]{
  \begin{minipage}[b]{.19\columnwidth}
    \includegraphics[width=0.99\textwidth]{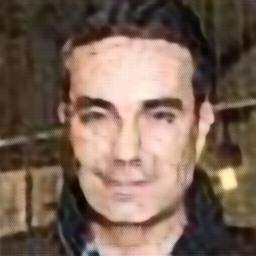}
  \end{minipage}
}
\hspace{-3ex}
\subfigure[\scriptsize Ours]{
  \begin{minipage}[b]{.19\columnwidth}
    \includegraphics[width=0.99\textwidth]{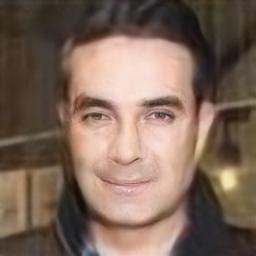}
  \end{minipage}
}
\vspace{-1ex}
\caption{Restoration on real low quality images compared with all the competing methods.}
\label{fig:realall}
\end{figure}
\begin{figure}[t]
\centering
\subfigure[]{
  \begin{minipage}[b]{.19\columnwidth}
    \includegraphics[width=0.99\textwidth]{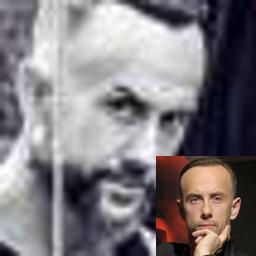}\\
    \includegraphics[width=0.99\textwidth]{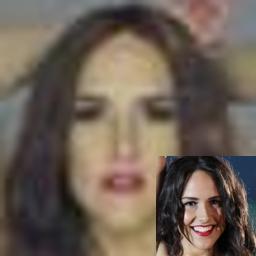}\\
    \includegraphics[width=0.99\textwidth]{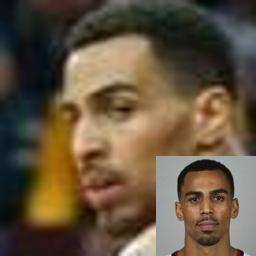}\\
    \includegraphics[width=0.99\textwidth]{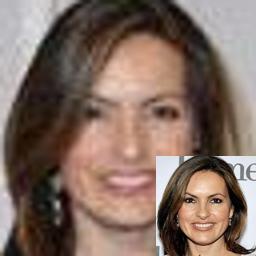}\\
    \includegraphics[width=0.99\textwidth]{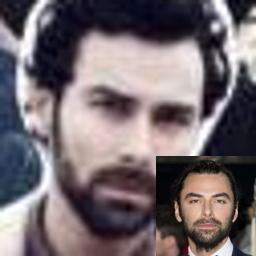}\\
    \includegraphics[width=0.99\textwidth]{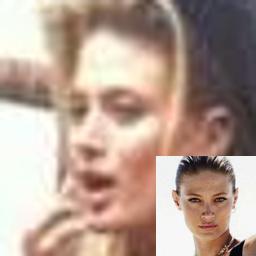}\\
    \includegraphics[width=0.99\textwidth]{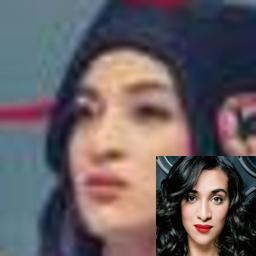}
  \end{minipage}
  \label{a}
}
\hspace{-3ex}
\subfigure[]{
  \begin{minipage}[b]{.19\columnwidth}
    \includegraphics[width=0.99\textwidth]{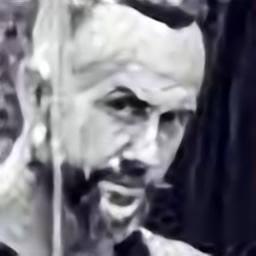}\\
    \includegraphics[width=0.99\textwidth]{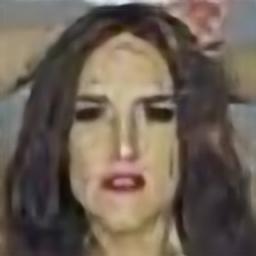}\\
    \includegraphics[width=0.99\textwidth]{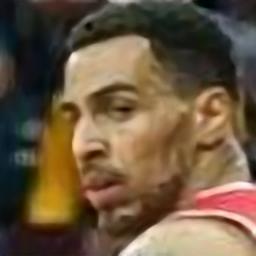}\\
    \includegraphics[width=0.99\textwidth]{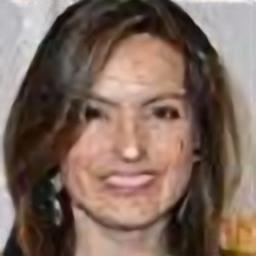}\\
     \includegraphics[width=0.99\textwidth]{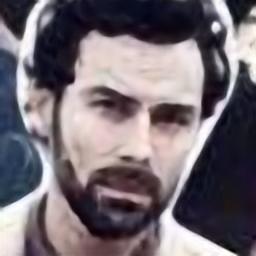}\\
    \includegraphics[width=0.99\textwidth]{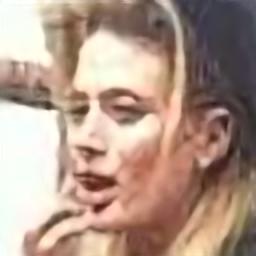}\\
    \includegraphics[width=0.99\textwidth]{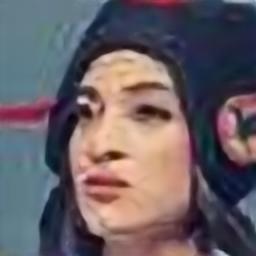}
  \end{minipage}
}
\hspace{-3ex}
\subfigure[]{
  \begin{minipage}[b]{.19\columnwidth}
    \includegraphics[width=0.99\textwidth]{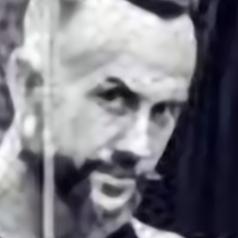}\\
    \includegraphics[width=0.99\textwidth]{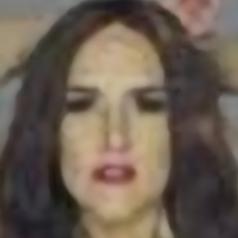}\\
    \includegraphics[width=0.99\textwidth]{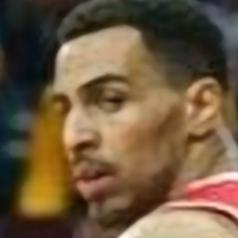}\\
    \includegraphics[width=0.99\textwidth]{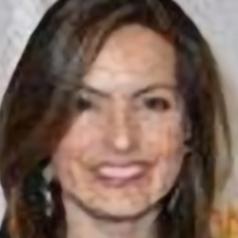}\\
    \includegraphics[width=0.99\textwidth]{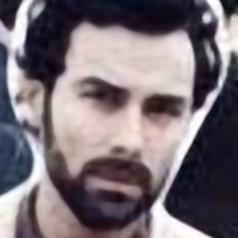}\\
    \includegraphics[width=0.99\textwidth]{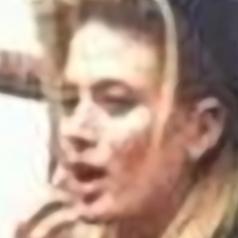}\\
    \includegraphics[width=0.99\textwidth]{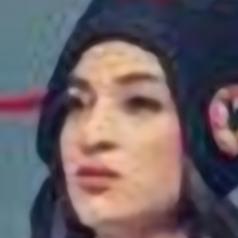}
  \end{minipage}
}
\hspace{-3ex}
\subfigure[]{
  \begin{minipage}[b]{.19\columnwidth}
    \includegraphics[width=0.99\textwidth]{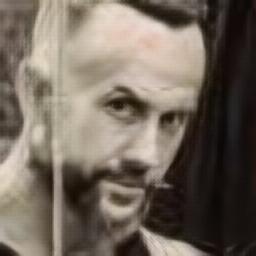}\\
    \includegraphics[width=0.99\textwidth]{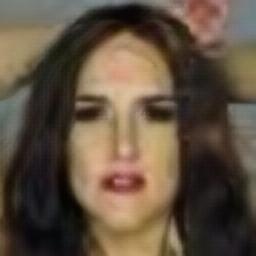}\\
    \includegraphics[width=0.99\textwidth]{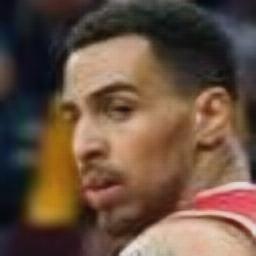}\\
    \includegraphics[width=0.99\textwidth]{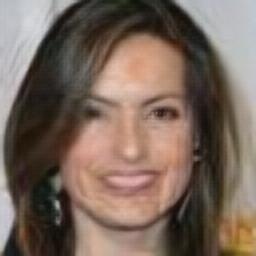}\\
    \includegraphics[width=0.99\textwidth]{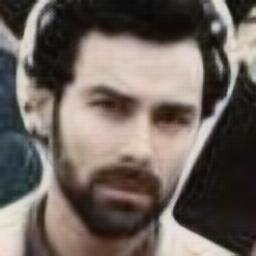}\\
    \includegraphics[width=0.99\textwidth]{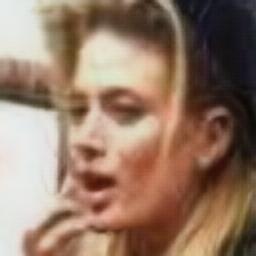}\\
    \includegraphics[width=0.99\textwidth]{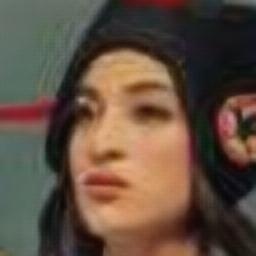}
  \end{minipage}
  \label{a}
}
\hspace{-3ex}
\subfigure[]{
  \begin{minipage}[b]{.19\columnwidth}
    \includegraphics[width=0.99\textwidth]{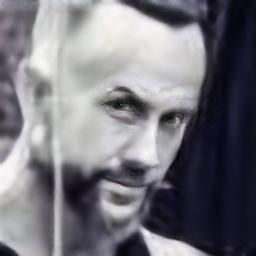}\\
    \includegraphics[width=0.99\textwidth]{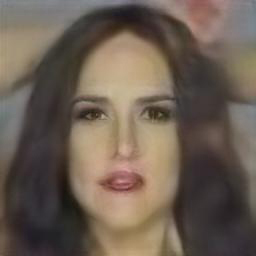}\\
    \includegraphics[width=0.99\textwidth]{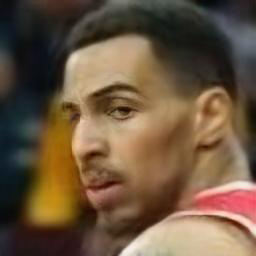}\\
    \includegraphics[width=0.99\textwidth]{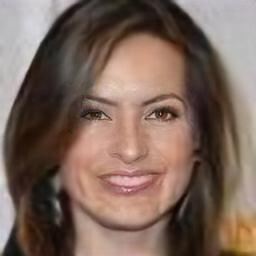}\\
    \includegraphics[width=0.99\textwidth]{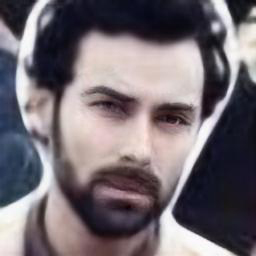}\\
    \includegraphics[width=0.99\textwidth]{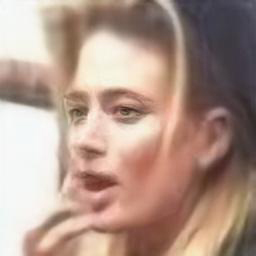}\\
    \includegraphics[width=0.99\textwidth]{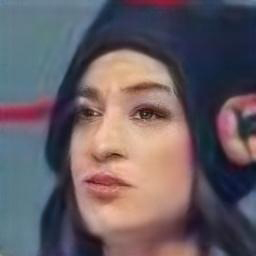}
  \end{minipage}
}
  \caption{Restoration results on real low quality images: (a)~real low quality images (Close-up in right bottom is the guided image), (b)~MDnCNN \cite{zhang2017beyond}, (c)~MARCNN \cite{Dong2015ICCV}, (d)~MDeblurGAN\cite{DeblurGAN}, and (e)~Ours.}
  \label{fig:real2}
\end{figure}
Fig.~\ref{fig:realall} shows the results on real low quality images by all the competing methods. As for pose problem, Fig.~\ref{fig:real2} shows more restoration results of our GFRNet compared with the top-3 performance methods on real low quality images with different poses. One can see that our GFRNet can also show great robustness in restoring facial images with different poses.
The real images are selected from VGGFace2 with the resolution lower than $60 \times 60$.
Even the degradation is unknown, our method yields visually realistic and pleasing results in face region with more fine details, while the competing methods can only achieve moderate improvement on visual quality.

\subsection{Ablation Studies}
\begin{figure}[t]
\centering
\subfigure[]{
  \begin{minipage}[b]{.19\columnwidth}
    \includegraphics[width=0.99\textwidth]{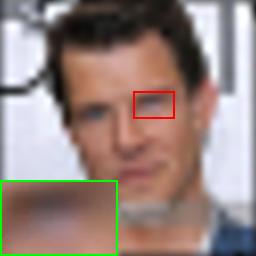}
  \end{minipage}
}
\hspace{-3ex}
\subfigure[]{
  \begin{minipage}[b]{.19\columnwidth}
    \includegraphics[width=0.99\textwidth]{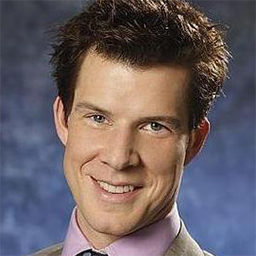}
  \end{minipage}
}
\hspace{-3ex}
\subfigure[]{
  \begin{minipage}[b]{.19\columnwidth}
    \includegraphics[width=0.99\textwidth]{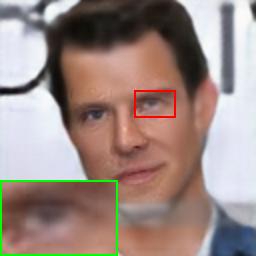}
  \end{minipage}
}
\hspace{-3ex}
\subfigure[]{
  \begin{minipage}[b]{.19\columnwidth}
    \includegraphics[width=0.99\textwidth]{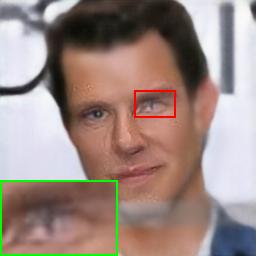}
  \end{minipage}
}
\hspace{-3ex}
\subfigure[]{
  \begin{minipage}[b]{.19\columnwidth}
    \includegraphics[width=0.99\textwidth]{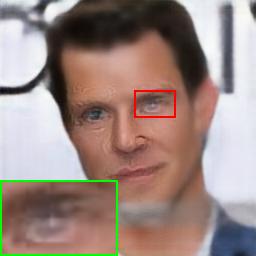}
  \end{minipage}
}
\hspace{-3ex}
\subfigure[]{
  \begin{minipage}[b]{.19\columnwidth}
    \includegraphics[width=0.99\textwidth]{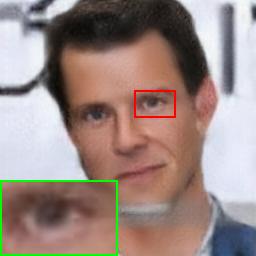}
  \end{minipage}
}
\hspace{-3ex}
\subfigure[]{
  \begin{minipage}[b]{.19\columnwidth}
    \includegraphics[width=0.99\textwidth]{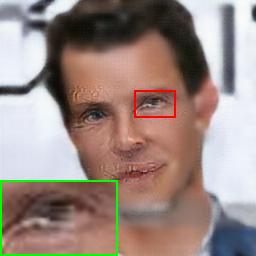}
  \end{minipage}
}
\hspace{-3ex}
\subfigure[]{
  \begin{minipage}[b]{.19\columnwidth}
    \includegraphics[width=0.99\textwidth]{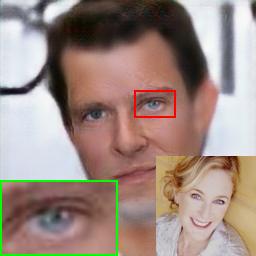}
  \end{minipage}
}
\hspace{-3ex}
\subfigure[]{
  \begin{minipage}[b]{.19\columnwidth}
    \includegraphics[width=0.99\textwidth]{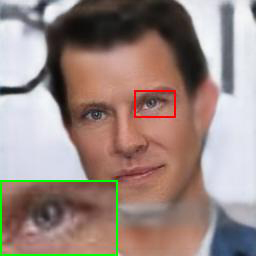}
  \end{minipage}
}
\hspace{-3ex}
\subfigure[]{
  \begin{minipage}[b]{.19\columnwidth}
    \includegraphics[width=0.99\textwidth]{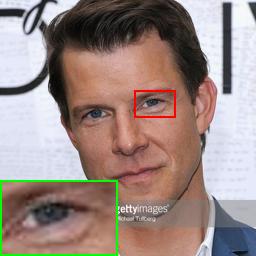}
  \end{minipage}
}
  \caption{Restoration results of our GFRNet variants: (a)~input, (b)~guided image. (c)~Ours($-WG$), (d)~Ours($-WG2$), (e)~Ours($-W$), (f)~Ours($-W2$), (g)~Ours($-F$), (h)~Ours($R$) (Close-up in right bottom is the random guided image), (i)~Ours($Full$), and (j)~ground-truth. Best viewed by zooming in the screen.}
  \label{fig:oursfff}
\end{figure}

\begin{figure}[t]
\centering
\subfigure[]{
  \begin{minipage}[b]{.19\columnwidth}
    \includegraphics[width=0.99\textwidth]{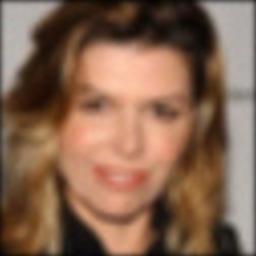}
  \end{minipage}
}
\hspace{-3ex}
\subfigure[]{
  \begin{minipage}[b]{.19\columnwidth}
    \includegraphics[width=0.99\textwidth]{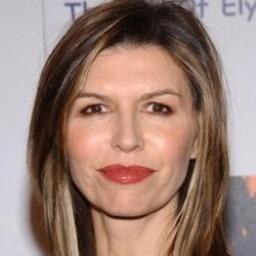}
  \end{minipage}
}
\hspace{-3ex}
\subfigure[]{
  \begin{minipage}[b]{.19\columnwidth}
    \includegraphics[width=0.99\textwidth]{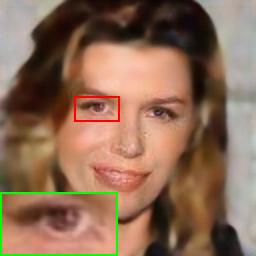}
  \end{minipage}
}
\hspace{-3ex}
\subfigure[]{
  \begin{minipage}[b]{.19\columnwidth}
    \includegraphics[width=0.99\textwidth]{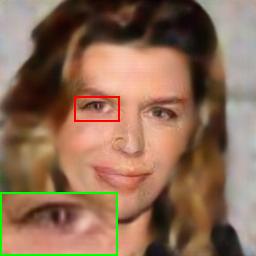}
  \end{minipage}
}
\hspace{-3ex}
\subfigure[]{
  \begin{minipage}[b]{.19\columnwidth}
    \includegraphics[width=0.99\textwidth]{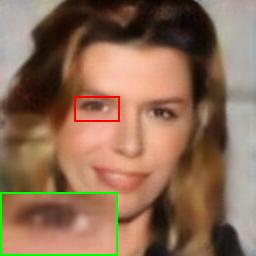}
  \end{minipage}
}
\hspace{-3ex}
\subfigure[]{
  \begin{minipage}[b]{.19\columnwidth}
    \includegraphics[width=0.99\textwidth]{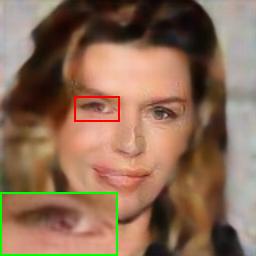}
  \end{minipage}
}
\hspace{-3ex}
\subfigure[]{
  \begin{minipage}[b]{.19\columnwidth}
    \includegraphics[width=0.99\textwidth]{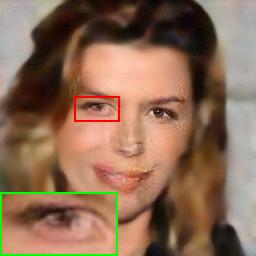}
  \end{minipage}
}
\hspace{-3ex}
\subfigure[]{
  \begin{minipage}[b]{.19\columnwidth}
    \includegraphics[width=0.99\textwidth]{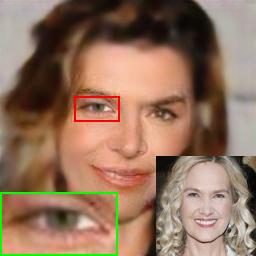}
  \end{minipage}
}
\hspace{-3ex}
\subfigure[]{
  \begin{minipage}[b]{.19\columnwidth}
    \includegraphics[width=0.99\textwidth]{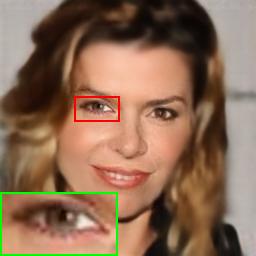}
  \end{minipage}
}
\hspace{-3ex}
\subfigure[]{
  \begin{minipage}[b]{.19\columnwidth}
    \includegraphics[width=0.99\textwidth]{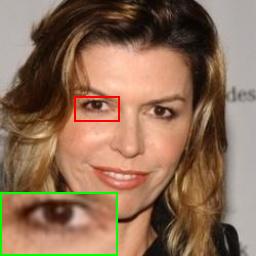}
  \end{minipage}
}
  \caption{Restoration results of our GFRNet variants: (a)~input, (b)~guided image. (c)~Ours($-WG$), (d)~Ours($-WG2$), (e)~Ours($-W$), (f)~Ours($-W2$), (g)~Ours($-F$), (h)~Ours($R$) (Close-up in right bottom is the random guided image), (i)~Ours($Full$), and (j)~ground-truth. Best viewed by zooming in the screen.}
  \label{fig:oursfff2}
\end{figure}

\begin{figure}[t]
\centering
\subfigure[]{
  \begin{minipage}[b]{.19\columnwidth}
    \includegraphics[width=0.99\textwidth]{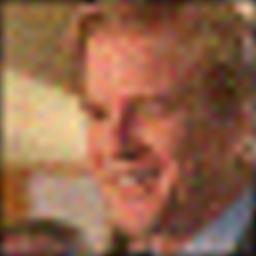}\\
    \includegraphics[width=0.99\textwidth]{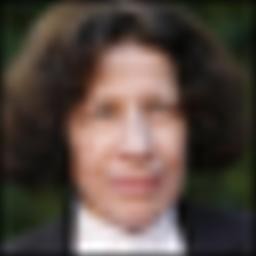}\\
    \includegraphics[width=0.99\textwidth]{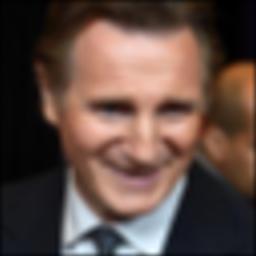}\\
    \includegraphics[width=0.99\textwidth]{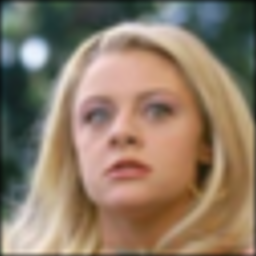}
  \end{minipage}
}
\hspace{-3ex}
\subfigure[]{
  \begin{minipage}[b]{.19\columnwidth}
    \includegraphics[width=0.99\textwidth]{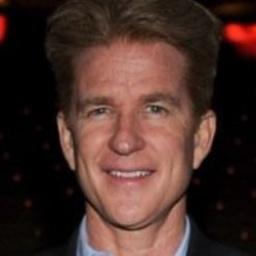}\\
    \includegraphics[width=0.99\textwidth]{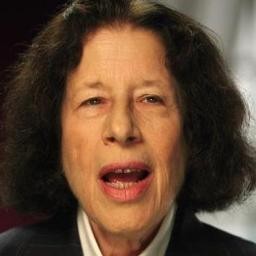}\\
    \includegraphics[width=0.99\textwidth]{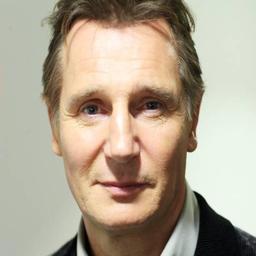}\\
    \includegraphics[width=0.99\textwidth]{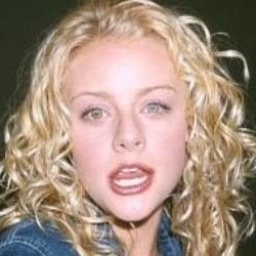}
  \end{minipage}
}
\hspace{-3ex}
\subfigure[]{
  \begin{minipage}[b]{.19\columnwidth}
    \includegraphics[width=0.99\textwidth]{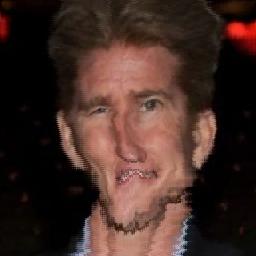}\\
    \includegraphics[width=0.99\textwidth]{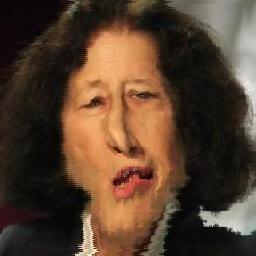}\\
    \includegraphics[width=0.99\textwidth]{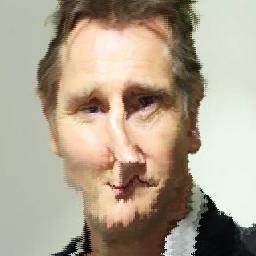}\\
    \includegraphics[width=0.99\textwidth]{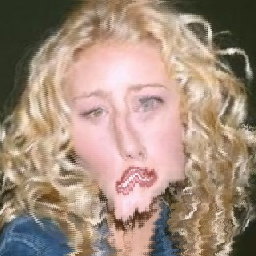}
  \end{minipage}
}
\hspace{-3ex}
\subfigure[]{
  \begin{minipage}[b]{.19\columnwidth}
    \includegraphics[width=0.99\textwidth]{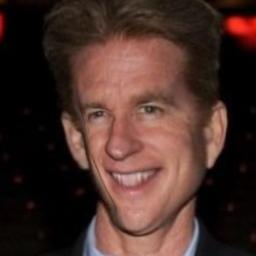}\\
    \includegraphics[width=0.99\textwidth]{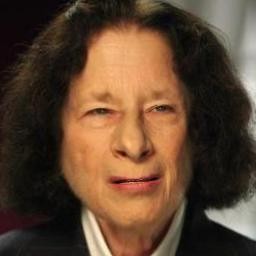}\\
    \includegraphics[width=0.99\textwidth]{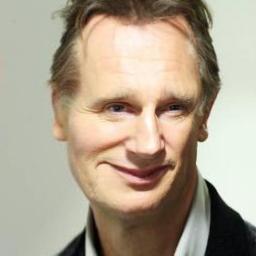}\\
    \includegraphics[width=0.99\textwidth]{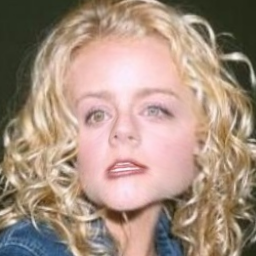}

  \end{minipage}
}
\hspace{-3ex}
\subfigure[]{
  \begin{minipage}[b]{.19\columnwidth}
    \includegraphics[width=0.99\textwidth]{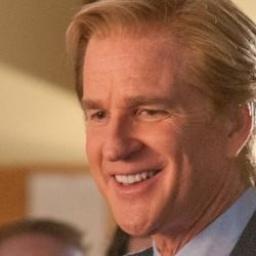}\\
    \includegraphics[width=0.99\textwidth]{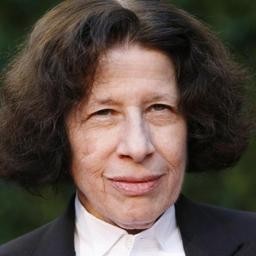}\\
    \includegraphics[width=0.99\textwidth]{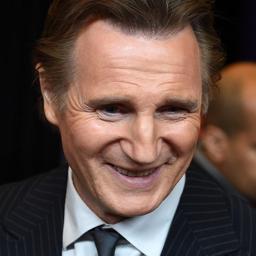}\\
    \includegraphics[width=0.99\textwidth]{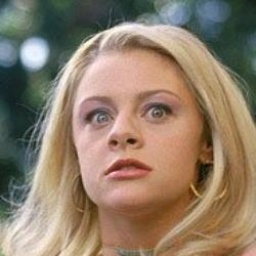}

  \end{minipage}
}
  \caption{{Warped guidance by Ours($Full$) and Ours($-F$)}: (a)~{input}, (b)~guided image, (c)~Ours($-F$), (d)~Ours($Full$), and (e)~ground-truth. }
  \label{fig:warp}

\end{figure}

Three groups of ablative experiments are conducted to assess the components of our GFRNet.
First, we consider five variants of our GFRNet:
(i) Ours($Full$): the full GFRNet,
(ii) Ours($-F$): GFRNet by removing the flow loss $\mathcal{L}_{flow}$,
(iii) Ours($-W$): GFRNet by removing WarpNet (RecNet takes both $I^d$ and $I^g$ as input),
(iv) Ours($-WG$): GFRNet by removing WarpNet and guided image (RecNet only takes $I^d$ as input),
and (v) Ours($R$): GFRNet by using a random $I^g$ with different identity to $I^d$.
Table~\ref{table::Quantization} also lists the PSNR and SSIM results of these variants, and we have the following observations.
(i) All the three components, i.e., guided image, WarpNet and flow loss, contribute to the performance improvement.
(ii) GFRNet cannot be well trained without the help of flow loss.
As a result, although Ours($-F$) outperforms Ours($-W$) in most cases, sometimes Ours($-W$) can perform slightly better than Ours($-F$) by average PSNR, e.g., for $8\times$ SR on VggFace2.
(iii) It is worth noting that GFRNet with random guidance (i.e., Ours($R$)) achieves the second best results, indicating that GFRNet is robust to the misuse of identity.
Figs.~\ref{fig:realfig1},~\ref{fig:oursfff} and~\ref{fig:oursfff2} give the restoration results by GFRNet variants.
Ours($Full$) can generate much sharper and richer details, and achieves better perceptual quality than its variants.
Moreover, Ours($R$) also achieves the second best performance in qualitative results, but it may introduce the fine details of the other identity to the result (e.g., eye regions in Fig.~\ref{fig:oursfff}(h)).
Furthermore, to illustrate the effectiveness of flow loss, Fig.~\ref{fig:warp} shows the warped guidance by Ours($Full$) and Ours($-F$).
Without the help of flow loss, Ours($-F$) cannot converge to stable solution and results in unreasonable warped guidance.
In contrast, Ours($Full$) can correctly align guided image to the desired pose and expression, indicating the necessity and effectiveness of flow loss.


\begin{figure}[t]
\centering
\subfigure[]{
  \begin{minipage}[b]{.19\columnwidth}
    \includegraphics[width=0.99\textwidth]{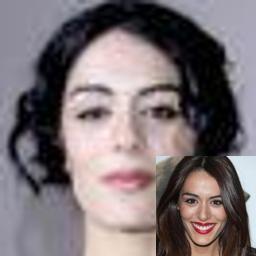}\\
    \includegraphics[width=0.99\textwidth]{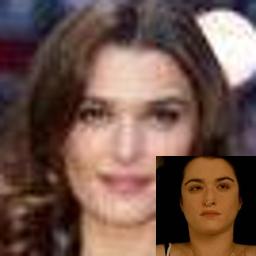}
  \end{minipage}
  \label{a}
}
\hspace{-3ex}
\subfigure[]{
  \begin{minipage}[b]{.19\columnwidth}
    \includegraphics[width=0.99\textwidth]{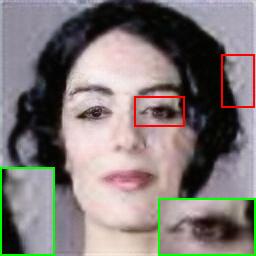}\\
    \includegraphics[width=0.99\textwidth]{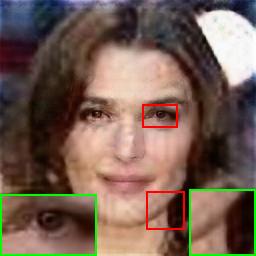}
  \end{minipage}
  \label{b}
}
\hspace{-3ex}
\subfigure[]{
  \begin{minipage}[b]{.19\columnwidth}
    \includegraphics[width=0.99\textwidth]{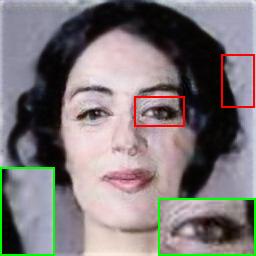}\\
    \includegraphics[width=0.99\textwidth]{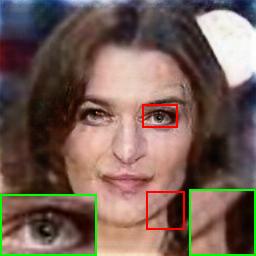}
  \end{minipage}
  \label{c}
}
\hspace{-3ex}
\subfigure[]{
  \begin{minipage}[b]{.19\columnwidth}
    \includegraphics[width=0.99\textwidth]{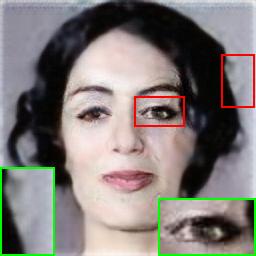}\\
    \includegraphics[width=0.99\textwidth]{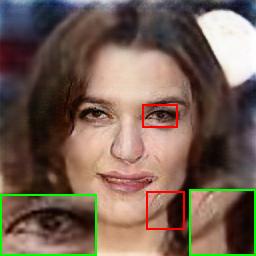}
  \end{minipage}
  \label{d}
}
\hspace{-3ex}
\subfigure[]{
  \begin{minipage}[b]{.19\columnwidth}
    \includegraphics[width=0.99\textwidth]{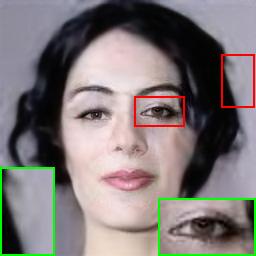}\\
    \includegraphics[width=0.99\textwidth]{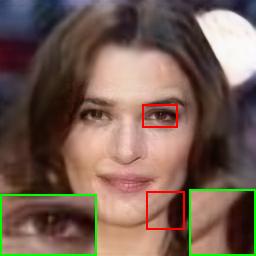}
  \end{minipage}
  \label{e}
}
  \caption{Results on real low quality images by our GFRNet trained with different degradation settings: (a)~real low quality image (Close-up in right bottom is the guided image), (b)~only blurring, (c)~blurring+downsampling, (d)~blurring+downsampling+AWGAN, and (e)~full degradation model. Best viewed by zooming in the screen.}
  \label{fig:blursamplenoisejpeg}
\end{figure}
Second, it is noted that the parameters of Ours($Full$) are nearly two times of Ours($-W$) and Ours($-WG$).
To show that the gain of Ours($Full$) does not come from the increase of parameter number, we include two other variants of GFRNet, i.e., Ours($-W2$) and Ours($-WG2$), by increasing the channels of Ours($-W$) and Ours($-WG$) to 2 times, respectively.
From Table~\ref{table::Quantization}, in terms of PSNR, Ours($Full$) also outperforms Ours($-W2$) and Ours($-WG2$) with a large margin.
Instead of the increase of model parameters, the performance improvement of Ours($Full$) should be mainly attributed to the incorporation of both WarpNet and flow loss.

Finally, four GFRNet models are trained based on four settings of our general degradation model: (i) only blurring, (ii) blurring+downsampling, (iii) blurring+downsampling+AWGN, (iv) our full degradation model in Eqn.~(\ref{eqn:degradation}).
Due to that the four models are trained using different degradation settings, it is unfair to compared them using any synthetic test data.
Thus, we test them on a real low quality image in Fig.~\ref{fig:blursamplenoisejpeg}.
It can be seen that the results by the settings (i)$\sim$(ii) are still blurry, while the results by the settings {(i)$\sim$(iii)} contain visual artifacts.
In comparison, the model by our full degradation model can produce sharp and clean result while suppressing most artifacts.
The results indicate that our full degradation model is effective in simulating real low quality images which usually have unknown and complex degradation.

\section{Conclusion}\label{section5}
In this paper, we present a guided blind face restoration model, i.e., GFRNet, by taking both the degraded observation and a high-quality guided image from the same identity as input.
Besides the reconstruction subnetwork, our GFRNet also includes a warping subnetwork (WarpNet), and incorporates the landmark loss as well as TV regularizer to align the guided image to the desired pose and expression.
To make our GFRNet be applicable to blind restoration, we further introduce a general image degradation model to synthesize realistic low quality face image.
Quantitative and qualitative results show that our GFRNet not only performs favorably against the relevant state-of-the-arts but also generates visually pleasing results on real low quality face images.
\clearpage
\bibliographystyle{splncs}
\bibliography{egbib}
\end{document}